\documentclass[10pt]{article} %
\usepackage[preprint]{tmlr}

\usepackage{amsmath,amsfonts,bm}

\def\eqref#1{equation~\ref{#1}}

\def\1{\bm{1}}

\DeclareMathAlphabet{\mathsfit}{\encodingdefault}{\sfdefault}{m}{sl}
\SetMathAlphabet{\mathsfit}{bold}{\encodingdefault}{\sfdefault}{bx}{n}

\usepackage{hyperref}
\usepackage{url}
\usepackage{graphicx}
\usepackage{listings}
\usepackage{color}
\usepackage{multirow}
\usepackage{array}
\usepackage{textgreek}
\usepackage{longtable}
\usepackage{caption}
\usepackage{booktabs}
\usepackage{subcaption}

\newcolumntype{C}[1]{>{\centering\let\newline\\\arraybackslash\hspace{0pt}}m{#1}}
\newcolumntype{L}[1]{>{\raggedright\let\newline\\\arraybackslash\hspace{0pt}}m{#1}}
\newcolumntype{R}[1]{>{\raggedleft\let\newline\\\arraybackslash\hspace{0pt}}m{#1}}

\definecolor{deepblue}{rgb}{0,0,0.7}
\definecolor{deepred}{rgb}{0.7,0,0}
\definecolor{deepgreen}{rgb}{0,0.5,0}

\newcommand\pythonstyle{\lstset{
    language=Python,
    basicstyle=\ttfamily,
    morekeywords={},              %
    keywordstyle=\ttfamily\color{deepblue},
    emph={get_n_params,get_flops_per_seq}, %
    emphstyle=\ttfamily\color{deepred},    %
    commentstyle=\color{deepgreen},
    stringstyle=\color{deepgreen},
    frame=tb,                         %
    showstringspaces=false
}}

\lstnewenvironment{python}[1][] {
    \pythonstyle
    \lstset{#1}
} {}

\makeatletter
\newcommand{\thickhline}{%
    \noalign {\ifnum 0=`}\fi \hrule height 1pt
    \futurelet \reserved@a \@xhline
}
\makeatother

\title{\centering Cerebras-GPT: Open Compute-Optimal Language Models Trained on the Cerebras Wafer-Scale Cluster}
\author{
    \centering
    \name Nolan Dey, Gurpreet Gosal, Zhiming (Charles) Chen, Hemant Khachane, William Marshall, \\ Ribhu Pathria, Marvin Tom, Joel Hestness \\ \vspace{6pt} %
    \affil Cerebras Systems~~~~
    \email \{nolan,joel\}@cerebras.net
}

\def\month{MM}  %
\def\year{YYYY} %
\def\openreview{\url{https://openreview.net/forum?id=XXXX}} %

\begin{document}

\maketitle

\begin{abstract}

We study recent research advances that improve large language models through efficient pre-training and scaling, and open datasets and tools.
We combine these advances to introduce Cerebras-GPT, a family of open compute-optimal language models scaled from 111M to 13B parameters.
We train Cerebras-GPT models on the Eleuther Pile dataset following DeepMind Chinchilla scaling rules for efficient pre-training (highest accuracy for a given compute budget).
We characterize the predictable power-law scaling and compare Cerebras-GPT with other publicly-available models to show all Cerebras-GPT models have state-of-the-art training efficiency on both pre-training and downstream objectives.
We describe our learnings including how Maximal Update Parameterization (\textmu P) can further improve large model scaling, improving accuracy and hyperparameter predictability at scale.
We release our pre-trained models and code, making this paper the first open and reproducible work comparing compute-optimal model scaling to models trained on fixed dataset sizes. Cerebras-GPT models are available on HuggingFace: \url{https://huggingface.co/cerebras}.

\end{abstract}

\section{Introduction}

Recent research in large language models (LLMs) shows important advances that can improve LLM quality and efficiency. Scaling law studies show predictable and significant improvements in model performance by increasing model and dataset size \citep{hestness2017scalinglaws,kaplan2020scalinglaws}. Language models can also be improved just by training on more data \citep{hoffmann2022chinchilla,llama}. Recent works, such as Maximal Update Parameterization (\textmu P), also show techniques to improve training stability and performance as models scale up (e.g., \cite{bachlechner2020rezero,yang2022mup}).

Concurrently with these advances, the research community has trained and released many open-source models. Models like GPT-J, GPT-NeoX, OPT, and Pythia have each held state-of-the-art accuracy for open source models for their size, and these models can be tested and used simply by downloading the pre-trained weights \citep{gpt-j,black2022gptneox,zhang2022opt,eleuther2023pythia}. While these models are important contributions, they have not aimed to be compute-efficient. The research community needs more reproducible scaling efforts that can guide collective decisions about training large foundation models in a compute-efficient way.

We introduce Cerebras-GPT, our open effort to combine recent LLM efficient scaling techniques to produce compute-optimal pre-trained models and corresponding scaling laws. Cerebras-GPT is a family of GPT-3-like models that we scale from 111M to 13B parameters. We train them on the open-source dataset, the Pile \citep{pile}, following DeepMind's Chinchilla scaling rules \citep{hoffmann2022chinchilla}. Cerebras-GPT models show state-of-the-art training efficiency when targeting both upstream Pile evaluations as well as a suite of downstream tasks. Our largest model shows state-of-the-art performance on pre-training and most downstream tasks compared to other comparably-sized public models. We also characterize some of the training stability challenges when scaling Cerebras-GPT. We address the challenges by training models with \textmu P, which shows further accuracy improvements and hyperparameter predictability.

\begin{figure}[h]
    \centering
    \includegraphics[width=0.7\linewidth]{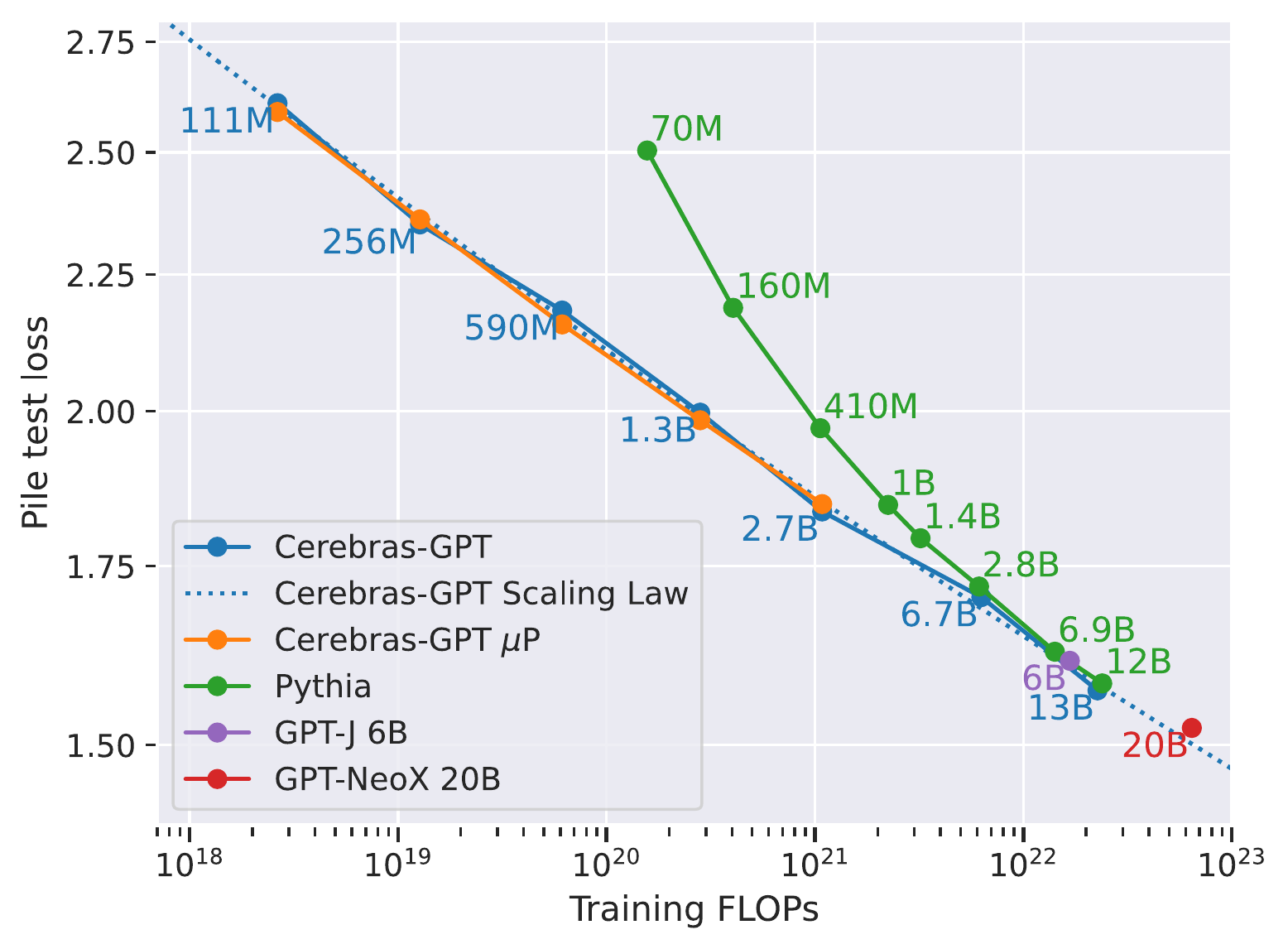}
    \vspace{-6pt}
    \caption{Pile test set loss given pre-training FLOPs for Cerebras-GPT, GPT-J, GPT-NeoX, and Pythia.}
    \label{figure:scaling_laws_front}
\end{figure}

Cerebras-GPT models form the compute-optimal Pareto frontier for both pre-training and popular downstream objectives. Figure \ref{figure:scaling_laws_front} shows the upstream Pile frontiers compared to contemporary works. We characterize the Pareto frontiers with scaling laws that can be used to predict the benefits of further model and dataset scaling efforts. We also observe and discuss that future open efforts should consider aggregate compute budget (both pre-training and expected inferences) when deciding the appropriate balance of model size and pre-training dataset size.

Overall, the contributions of this work are as follows:
\begin{itemize}
    \vspace{-6pt}
    \setlength\itemsep{0px}
    \item We train Cerebras-GPT compute-optimal models scaled from 111M to 13B parameters on the Pile dataset following Chinchilla scaling rules to collect compute-efficient scaling laws.
    \item We show that these models provide state-of-the-art pre-training efficiency on both pre-training and downstream objectives compared to other open models--the first such open effort.
    \item We provide detailed instructions to reproduce our results, including the use of \textmu P to improve training stability and transfer hyperparameters as models scale up.
    \item We document our experience training these models on the Andromeda AI Cluster, comprising 16 Cerebras CS-2 systems, and we describe the simplicity of scaling models and performance.
\end{itemize}

Finally, we aim to enable the research community to consume these results. We release our pre-trained models and code, and we share details about our training process here, so the community can use and reproduce our results. Pre-trained models are available on HuggingFace: \url{https://huggingface.co/cerebras}. Source code is available in the Cerebras Modelzoo: \url{https://github.com/Cerebras/modelzoo}. We hope these models will be a valuable addition for the open-source community.

\section{Methodology}

In this section, we describe the details of the models we trained, including hyperparameters used at each model scale and details about how we obtain and use the Pile dataset. We also motivate the need for techniques to stabilize scaling, and we describe how we use Maximal Update Parameterization (\textmu P).

\subsection{Model Architecture}
\label{section:architecture}
Cerebras-GPT models have a GPT-3-like architecture, an autoregressive transformer decoder model \citep{gpt3}. The main difference is that unlike GPT-3, which uses alternating dense and sparse-banded attention, we use dense attention in all decoder blocks. We select model dimensions to either follow aspect ratio ${\sim}80$ ($d_{\text{model}}/n_{\text{layers}}$) or the same shape as GPT-3 models. All models are trained with a maximum sequence length of 2048 tokens. Table \ref{table:model_details} lists the specific model dimensions for each model size. Our formula for the number of parameters is provided in appendix \ref{appendix:n_params}.

\begin{table}[h]
    \centering
    \caption{Cerebras-GPT model architecture and training algorithm details}
    \vspace{-6pt}
    \label{table:model_details}
    \begin{tabular}{l|rrrr|rccc}
                   & \multicolumn{4}{c|}{Model Dimensions} & \multicolumn{1}{c}{Total} & \multicolumn{1}{c}{Batch Size} & \multicolumn{1}{c}{Learning} & LR Decay \\
        Parameters & \multicolumn{1}{c}{$d_{\text{model}}$} & \multicolumn{1}{c}{$n_{\text{layers}}$} & \multicolumn{1}{c}{$d_{\text{head}}$} & \multicolumn{1}{c|}{$d_{\text{ffn}}$} & \multicolumn{1}{c}{tokens} & \multicolumn{1}{c}{(tokens)} & \multicolumn{1}{c}{Rate (LR)} & Type \\
        \hline
        111M       & 768      & 10        & 64      & 3072   &   2.2B & 246K              & 6.0E-04  & Linear   \\
        256M       & 1088     & 14        & 64      & 4352   &   5.1B & 541K              & 6.0E-04  & Linear   \\
        590M       & 1536     & 18        & 128     & 6144   &  11.8B & 541K              & 2.0E-04  & Linear   \\
        1.3B       & 2048     & 24        & 128     & 8192   &  26.3B & 1.08M             & 2.0E-04  & Cosine   \\
        2.7B       & 2560     & 32        & 80      & 10240  &  53.0B & 1.08M             & 2.0E-04  & Cosine   \\
        6.7B       & 4096     & 32        & 128     & 16384  & 133.2B & 2.13M             & 1.2E-04  & Linear   \\
        13B        & 5120     & 40        & 128     & 20480  & 257.1B & 1.47M$\to$2.21M   & 1.2E-04  & Cosine   \\
        \hline
        111M + \textmu P   & 768      & 10        & 64      & 3072   &   2.2B & 246K      & 6.0E-03  & Linear   \\
        256M + \textmu P   & 1088     & 14        & 64      & 4352   &   5.1B & 541K      & 6.0E-03  & Linear   \\
        590M + \textmu P   & 1536     & 18        & 128     & 6144   &  11.8B & 541K      & 6.0E-03  & Linear   \\
        1.3B + \textmu P   & 2048     & 24        & 128     & 8192   &  26.3B & 1.08M     & 6.0E-03  & Linear   \\
        2.7B + \textmu P   & 2560     & 32        & 80      & 10240  &  53.0B & 1.08M     & 6.0E-03  & Linear   \\
        \end{tabular}
\end{table}

\subsection{Pre-training Corpus}
\label{section-pretraining-corpus}
We pre-train models on the Pile dataset, which consists of data from 22 data sources, including Common Crawl, PubMed Central, Books3, OpenWebText2, Github, and arXiv \citep{pile}. We use the dataset splits for train, test, and validation sets provided in the Pile configuration. We tokenize the corpora with byte-pair encoding and the GPT-2 vocabulary of size 50257 \citep{sennrich-etal-2016-neural,radford2019gpt2}. We do not perform deduplication of Pile but believe that deduplication could further improve our results. We include more details about Pile and dataset pre-processing in Appendix \ref{appendix_dataset}.

To evaluate pre-training, we compare Cerebras-GPT models to several publicly-available models using cross-entropy loss on the Pile test set. To ensure fair comparisons, we run evaluation ourselves on all checkpoints rather than using published numbers, though in most cases, our evaluations match prior works. For models that use different vocabularies, we correct cross-entropy back to the equivalent value with the GPT-2 vocabulary based on the number of tokens in each dataset.

\subsection{Model Training}
\label{section:training-setup}
We train models using the following training configurations. We use the AdamW optimizer \citep{AdamW} with (beta1, beta2) = ($0.9$, $0.95$). We set epsilon to $1e$-$8$ for small models and to $1e$-$9$ for 6.7B and 13B parameter models. We use weight decay of $0.1$ for all models. We do not use dropout for pre-training. For all runs, we use gradient norm clipping of $1.0$.

We use learning rates and batch sizes consistent with prior works, as listed in Table \ref{table:model_details}. We find that linear learning rate decay tends to perform better than cosine decay, so we use it in most of our pre-training runs. With either decay type, we warm up learning rate linearly over 375M tokens and then decay to $10\%$ of the maximum learning rate. The table also includes batch sizing. For the 13B parameter model, we train with a batch size of 720 sequences of length 2048 tokens for the first 84B tokens. At that point, we observed the gap between validation and train loss growing, indicating that the gradient noise was growing, so we increased the batch size to 1080 sequences for the rest of training.

To scale Cerebras-GPT model training in a compute-efficient way, we follow the DeepMind Chinchilla scaling methodology outlined in \citep{hoffmann2022chinchilla}. Specifically, we test and find that models trained with roughly 20 tokens per parameter offer the most compute-efficient pre-training, consistent with the Chinchilla results. We believe this paper is the first open effort to estimate the compute-efficient tokens per parameter for the Pile dataset. Our results Section \ref{section:pre-training_results} characterizes the effect of training with more tokens per parameter, and we include further test results in Appendix \ref{appendix:tokens_per_parameter}.

Finally, we train models using both FP16 mixed precision and bfloat16 precision \citep{micikevicius2018mixedprecision,abadi2015tfbfloat16}. Overall, we find bfloat16 to be more stable due to its extra exponent range, so we use it for all Cerebras-GPT models that we release. We include further discussion of precision in Appendix \ref{appendix:stable_training}.

\subsection{Standard (SP) and Maximal Update Parameterization (\textmu P)}
\label{section:mup_method}

{\bf Standard Parameterization (SP)}: We configure our main Cerebras-GPT models with the common standard parameterization (SP) approach. In SP, model weights are initialized from normal distributions with constant standard deviation or standard deviation based on the shape of each layer \citep{glorot2010initialization}. We initialize embedding and hidden layer weights with a truncated normal distribution with standard deviation $\sigma = 0.02$. An exception is that we use a standard deviation of $\sigma = 0.02/\sqrt{2 \cdot n_{\text{layers}}}$ for the last layer inside each residual network, following the GPT-2 initialization \citep{radford2019gpt2}.

Unfortunately, the SP approach does not account for potential inter-layer interactions and resulting training dynamics that arise when scaling to very large models. As SP models scale, they tend to become unstable as weight and activation values bump up against the limits of the numerical representations used to train them. For large models, unstable training can cause very costly restarts and researchers might not have budget for extensive hyperparameter tuning.

{\bf Maximal Update Parameterization (\textmu P)}: To address these issues, we also train a set of Cerebras-GPT models with Maximal Update Parameterization (\textmu P) \citep{yang2022mup}. \textmu P controls initialization, layer-wise learning rates, and activation magnitudes to ensure analytically stable training independent of a model's layer widths. In addition to improving training stability, \textmu P also improves the transferability of training hyperparameters from smaller to larger scale models, a technique called \textmu Transfer. \textmu Transfer permits directly using the same settings for some optimizer hyperparameters, most notably the learning rate.

We train a set of Cerebras-GPT models using \textmu P. We follow the \textmu Transfer approach by first tuning hyperparameters for a small, 40M parameter \textmu P model. Then, we transfer the hyperparameters along our \textmu P scaling law up to a 2.7B parameter model. \textmu P requires small changes to our baseline Cerebras-GPT models, including adding element-wise activation tensor scaling, adjusting initializers for affected layers, and adding layer-wise learning rates scaling to certain layers. We discuss the benefits we see with \textmu P in Section \ref{section:mup_results}. Refer to Appendix \ref{appendix:mup_implementation} for our tips to implement \textmu P and our hyperparameter tuning notes.

\section{Results}\label{section-results}

In this section, we show pre-training and downstream evaluations of Cerebras-GPT models, scaled from 111M to 13B parameters, and we compare against recent related works. We characterize the compute-efficient Pareto frontier for pre-training models on the Pile dataset and show that models on this frontier are also competitive on downstream tasks. We believe this is the first study to release a compute-optimal scaling law for pre-training on the Pile dataset that is openly reproducible by the community.

We show that Cerebras-GPT models define the state-of-the-art compute-optimal Pareto frontier on both pre-training and downstream objectives. Further, our largest model with 13B parameters shows improved accuracy on most downstream tasks compared to other comparably-sized publicly-available models\footnote{We believe the LLaMa 13B model is better than Cerebras-GPT on downstream tasks because it was trained for $4\times$ more tokens, but were unable to get access to test the model ourselves.}. We also train Cerebras-GPT models configured using \textmu P. We show that \textmu P enables direct hyperparameter transfer from smaller to larger models and improves the compute-optimal frontier loss by $0.4\%$.

\subsection{Pre-training Results}
\label{section:pre-training_results}
We scaled and pre-trained Cerebras-GPT models from 111M--13B parameters on the Pile dataset. We compare the Pile test set loss\footnote{All Cerebras-GPT development and hyperparameter tuning was evaluated using the Pile validation set.} for Cerebras-GPT models against other publicly available pre-trained models, GPT-J, GPT-NeoX, and Pythia \citep{gpt-j,black2022gptneox,eleuther2023pythia}. We believe these models to be fair comparisons either because the models were trained directly on Pile or on similarly-prepared datasets.

Figure \ref{figure:scaling_laws} plots pre-training efficiency (values also listed in Table \ref{table:large_model_evals}). The horizontal axis plots floating-point operations (FLOPs) spent during pre-training (log scale), and the vertical axis plots Pile test loss (log scale)\footnote{Pile test loss is crossentropy in nats/token. We correct all crossentropy results for different vocabularies to be comparable to the GPT-2 vocabulary.}. Across all model scales, Cerebras-GPT sets the efficiency frontier, largely because models were pre-trained with 20 tokens per parameter, consistent with findings in the Chinchilla paper. Other public models use more tokens per parameter, requiring more FLOPs to achieve similar loss. 

\begin{table}[h]
    \centering
    \begin{minipage}{0.49\linewidth}
        \centering
        \includegraphics[width=0.95\linewidth]{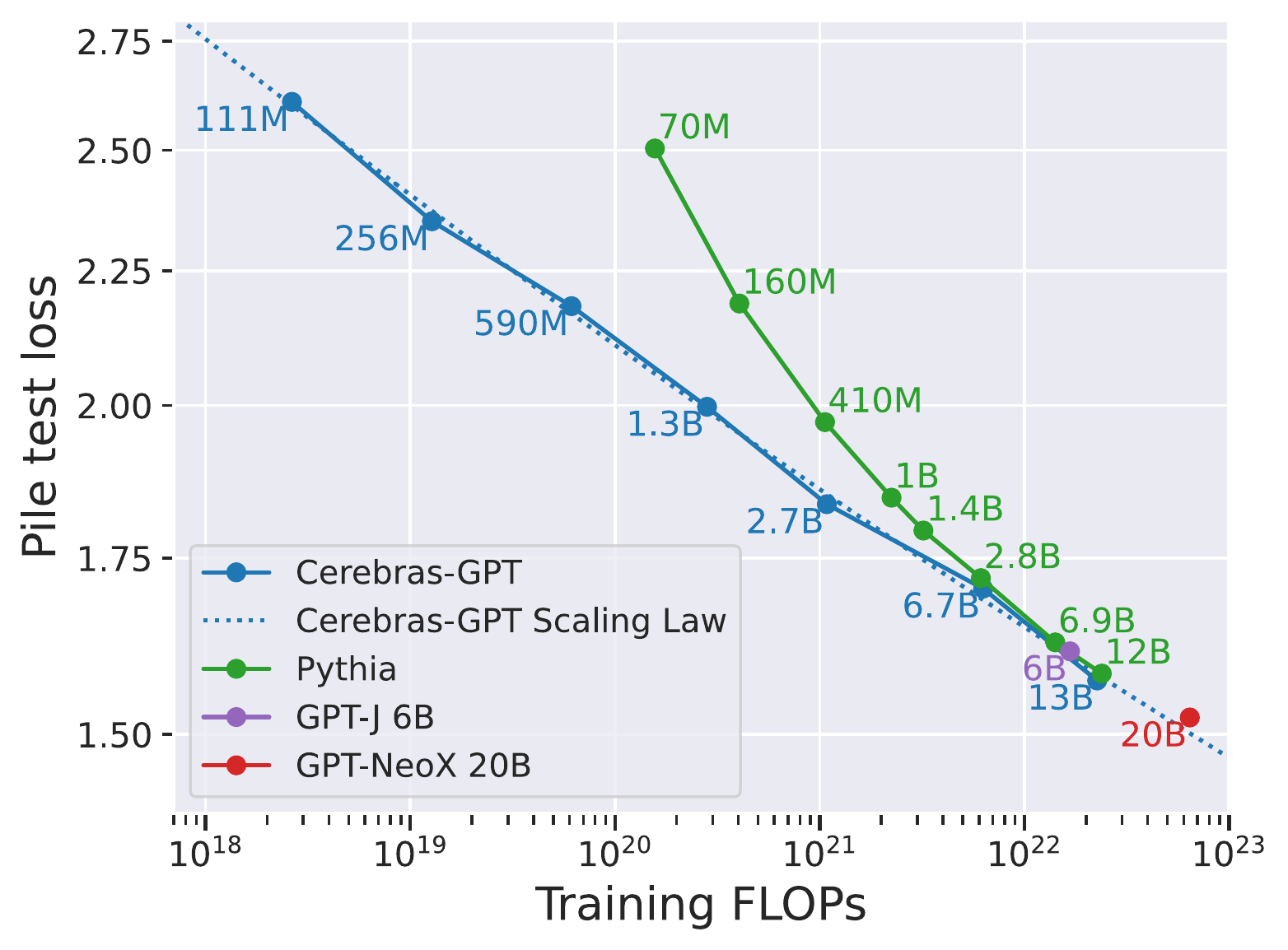}
        \vspace{-6pt}
        \captionof{figure}{\centering Pile test set loss given pre-training FLOPs for Cerebras-GPT, GPT-J, GPT-NeoX, and Pythia.}
        \label{figure:scaling_laws}
    \end{minipage}
    \begin{minipage}{0.49\linewidth}
	\centering
        \includegraphics[width=0.95\linewidth]{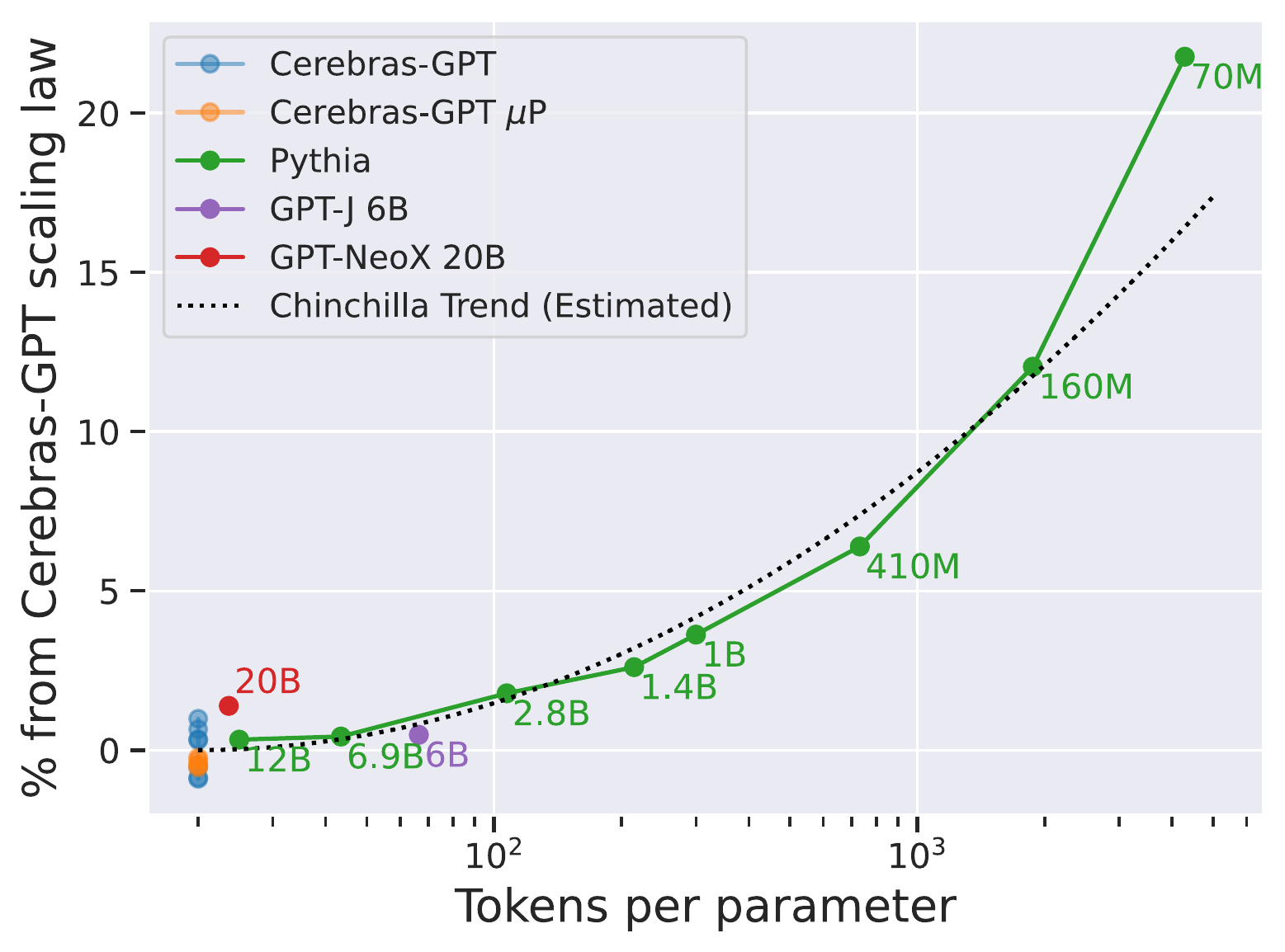}
        \vspace{-6pt}
	\captionof{figure}{\centering Percent loss degradation from Cerebras-GPT compute-optimal scaling law.}
	\label{figure:tpp}
    \end{minipage}
\end{table}

There are a couple notable observations from Figure \ref{figure:scaling_laws}. First, the scaling law for Cerebras-GPT models extrapolates accurately to larger model scales. We estimated the 13B model loss using a similar scaling law from models up to 6.7B parameters, and the 13B model trained to within 0.5\% of projected loss. Extending the existing scaling law shows that if we budgeted to train a model with FLOPs equivalent to GPT-NeoX 20B, we would expect the Cerebras-GPT model loss to be ${\sim}1.2\%$ better than GPT-NeoX 20B. For future reference, we include the compute-optimal frontier scaling law here ($f$ is compute FLOPs to loss, $\mathcal{L}$):

\vspace{-6pt}
\begin{equation}
    \mathcal{L}(f) = (f/5.984e22)^{-0.0737} + 0.5066
\end{equation}

Second, increasing tokens per parameter above 20 leads to smoothly degraded loss for the FLOP budget. Pythia models are each trained using 299.9B tokens from the Pile. As model size increases, tokens per parameter decreases reciprocally, and losses move closer to the compute-optimal frontier. The largest Pythia model at 12B parameters is trained with 25.3 tokens per parameter and is just 0.3\% loss above the Cerebras-GPT scaling law.

The loss gap from the compute-optimal frontier appears to be predictable in terms of tokens per parameter. In Figure \ref{figure:tpp}, we plot the percentage loss increase compared to the Cerebras-GPT frontier as a function of tokens per parameter. Here, Cerebras-GPT models cluster at 20 tokens per parameter, and Pythia results show the smooth curve away from the frontier for more tokens per parameter. We also include an estimate of the Chinchilla loss degradation from curve fitting data in their plots (\cite{hoffmann2022chinchilla}, Figure 3). These results confirm the estimate that compute optimal pre-training on the Pile should use roughly 20 tokens per parameter, a striking consistency with the Chinchilla results on the MassiveText dataset. Further tokens per parameter tests are in Appendix \ref{appendix:tokens_per_parameter}.

\subsection{Downstream Results}

We evaluate Cerebras-GPT and publicly-available models on a suite of seven common sense reasoning tasks in both the zero-shot and five-shot settings using the EleutherAI evaluation harness \citep{eval-harness}. In particular, we evaluate models on the tasks HellaSwag, PIQA, WinoGrande, Lambada, ARC (both the easy and challenge versions), and OpenBookQA \citep{hellaswag,piqa,winogrande,lambada,arc,openbookqa}. We include more detail about these tasks in Appendix \ref{section:downstream_tasks}. In addition to models we evaluate on upstream Pile, we add downstream results for OPT models \citep{zhang2022opt}, which were trained on a broader dataset but still using 300B pre-training tokens.

\begin{figure}[h]
    \centering
    \includegraphics[width=0.9\linewidth]{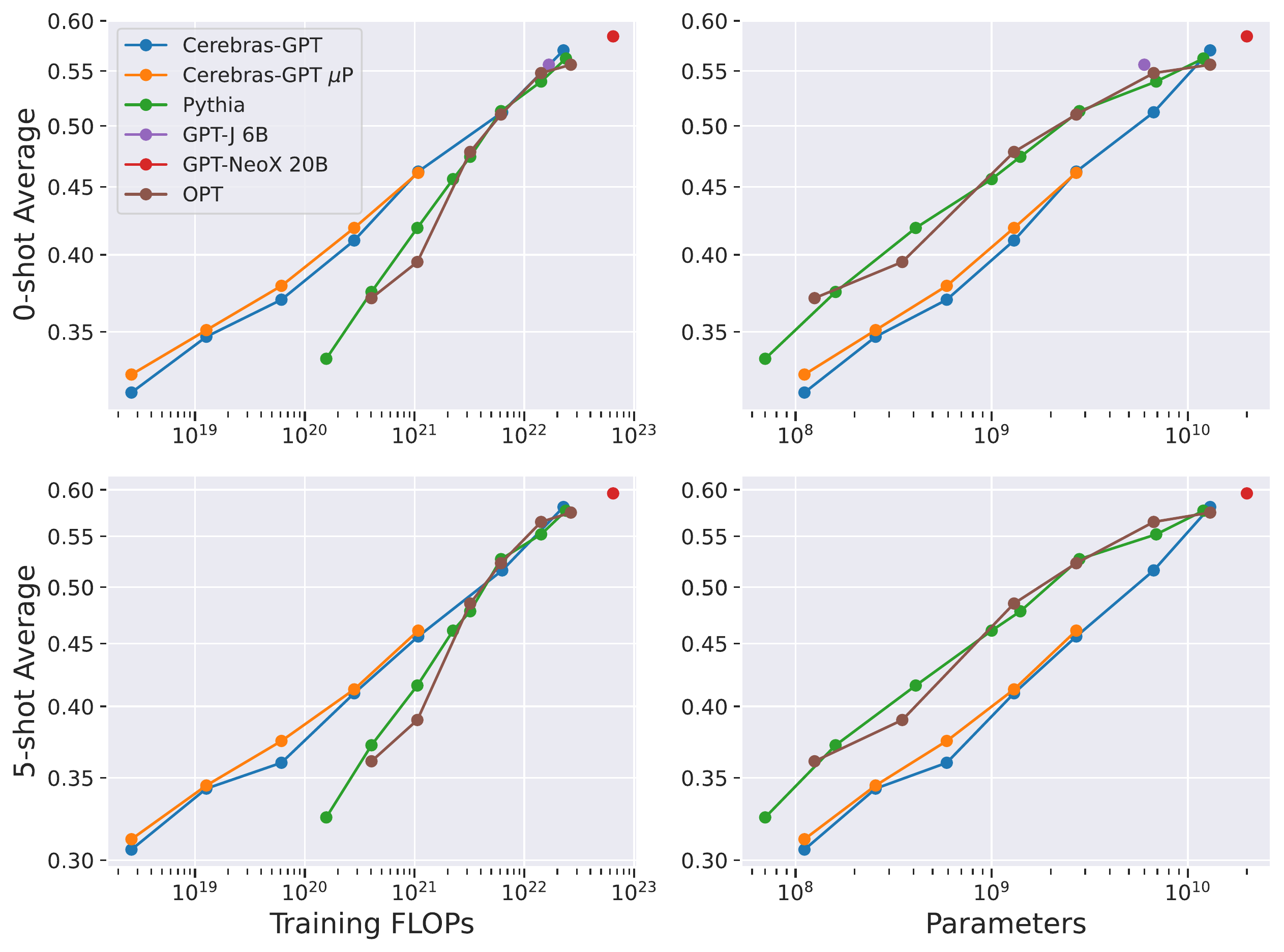}
    \vspace{-6pt}
    \caption{\centering Average zero- and five-shot downstream task accuracy plotted against FLOPs (left) and parameters (right). Higher accuracy is better. Individual tasks are plotted in Figures \ref{figure:downstream_all_flops} and \ref{figure:downstream_all_params}.}
    \label{figure:downstream_results}
\end{figure}

Like the pre-training results, Cerebras-GPT models form the compute-optimal Pareto frontier for downstream tasks as well. Figure \ref{figure:downstream_results} summarizes the average downstream task results for both zero- and five-shot evaluations\footnote{Here, we report accuracy result from each model predictions using token-level probability, consistent with reported results in the GPT-NeoX paper. We report additional accuracy measures in Appendix \ref{appendix:downstream_tasks}.} comparing Cerebras-GPT to GPT-J, GPT-NeoX, and Pythia. As Pythia and OPT models grow close to the 20 tokens per parameter count, they approach the Cerebras-GPT frontier FLOPs-to-accuracy. Here again, the Cerebras-GPT 13B model shows the best average downstream result for models of comparable size.

Figure \ref{figure:downstream_results} also plots downstream averages against model size in parameters (right column). For each model size smaller than 13B parameters, GPT-J, OPT, and Pythia models show significantly better downstream accuracy than Cerebras-GPT models, as expected. The Pythia and OPT accuracy frontiers deflect from straight lines (power-laws in log-log-scale), whereas Cerebras-GPT frontiers continue, indicating that downstream accuracy is predictable by model size for models trained with fixed tokens-per-parameter. The Cerebras-GPT trend suggests these models would be competitive with GPT-NeoX 20B if scaled to that size.

\begin{table}[h]
\caption{Zero-shot downstream task results for large publicly-available models. Full results in Table \ref{table:eval_results_zero_shot}.}
\vspace{-6pt}
\label{table:large_model_evals}
\resizebox{\textwidth}{!}
{
\begin{tabular}{lc|C{1.3cm}C{1.4cm}|C{1.0cm}cC{1.0cm}C{1.0cm}ccC{1.0cm}C{1.1cm}}
    \thickhline
        &    & \multicolumn{2}{c|}{Pre-training ($\downarrow$)} & \multicolumn{8}{c}{Downstream task accuracy ($\uparrow$)} \\
    \multicolumn{2}{c|}{Model} & Training FLOPs & Pile test xent & Hella- Swag &     PIQA & Wino- Grande &  Lambada &    ARC-e &    ARC-c & OpenBookQA & Downstream Avg. \\ \hline
    OPT                & 2.7B &        6.1e21 &           - & {\bf 0.458} & {\bf 0.738} &       0.610 &       0.637 &       0.609 &       0.268 & {\bf 0.250} &       0.510 \\
    Pythia             & 2.8B &        6.1e21 & {\bf 1.720} &       0.451 &       0.737 & {\bf 0.612} & {\bf 0.654} & {\bf 0.629} & {\bf 0.288} &       0.220 & {\bf 0.513} \\
    Cerebras-GPT       & 2.7B &  {\bf 1.1e21} &       1.834 &       0.386 &       0.701 &       0.559 &       0.567 &       0.571 &       0.246 &       0.206 &       0.462 \\
    \hline
    GPT-J              & 6.1B &        1.7e22 & {\bf 1.613} & {\bf 0.518} &       0.752 &       0.640 & {\bf 0.683} & {\bf 0.670} & {\bf 0.340} & {\bf 0.288} & {\bf 0.556} \\
    OPT                & 6.7B &        1.4e22 &           - &       0.505 & {\bf 0.763} & {\bf 0.654} &       0.677 &       0.656 &       0.307 &       0.276 &       0.548 \\
    Pythia             & 6.9B &        1.4e22 &       1.626 &       0.482 &       0.746 &       0.611 &       0.679 &       0.669 &       0.323 &       0.270 &       0.540 \\
    Cerebras-GPT       & 6.7B &  {\bf 6.3e21} &       1.704 &       0.447 &       0.739 &       0.602 &       0.636 &       0.643 &       0.282 &       0.238 &       0.512 \\
    \hline
    OPT                &  13B &  2.7e22       &           - & {\bf 0.524} &       0.759 & {\bf 0.651} &       0.687 &       0.671 &       0.329 &       0.270 &       0.556 \\
    Pythia             &  12B &  2.4e22       &       1.582 &       0.505 &       0.761 &       0.645 & {\bf 0.705} &       0.700 &       0.336 &       0.284 &       0.562 \\
    Cerebras-GPT       &  13B &  {\bf 2.3e22} & {\bf 1.572} &       0.513 & {\bf 0.766} &       0.646 &       0.696 & {\bf 0.714} & {\bf 0.367} & {\bf 0.286} & {\bf 0.570} \\
    \hline
    GPT-NeoX           &  20B &  {\bf 6.4e22} & {\bf 1.519} & {\bf 0.535} & {\bf 0.779} & {\bf 0.661} & {\bf 0.720} & {\bf 0.723} & {\bf 0.380} & {\bf 0.290} & {\bf 0.584} \\
    \hline
    \hline
    \multirow{3}{*}{\begin{tabular}[x]{@{}l@{}}Pythia\\Pile-dedup\end{tabular}} & 2.8B &        6.1e21 &       1.724 &       0.466 &       0.743 &       0.612 &       0.672 &       0.662 &       0.299 &       0.232 &       0.526 \\
                       & 6.9B &        1.4e22 &       1.644 &       0.488 &       0.756 &       0.636 &       0.695 &       0.667 &       0.320 &       0.252 &       0.545 \\
                       &  12B &        2.4e22 &       1.601 &       0.516 &       0.761 &       0.639 &       0.712 &       0.697 &       0.341 &       0.280 &       0.564 \\
    \thickhline
\end{tabular}
}
\end{table}

Finally, Table \ref{table:large_model_evals} shows more detailed downstream task comparisons for large publicly-available models, grouped into comparable sizes. We bold the results that are the best for each task and model size group. Each model family has at least one model that is best for some tasks. In this table, we also include results for Pythia models trained on a deduplicated version of the Pile. We separated these results, since they may not be directly comparable to others above, which were trained using the same or similar dataset preparation. As expected from the deduplication process, Pythia models show more difficulty generalizing to the pre-training Pile test loss task than other open models, which might have seen duplicated data during training. However, the Pythia Pile-dedup models typically improve accuracy on downstream tasks (1.8\% on average), indicating the potential benefits of deduplication.

\subsection{Maximal Update Parameterization (\textmu P) and \textmu Transfer}
\label{section:mup_results}

As we scaled the Cerebras-GPT models with standard parameterization (SP) along our scaling law, we experienced challenges predicting appropriate hyperparameters, and these models show substantial variance around their common scaling law. To address these challenges, we also test \textmu P and \textmu Transfer tuned hyperparameters to 111M--2.7B parameter Cerebras-GPT models. Across model sizes, our \textmu P models exhibit an average of 0.43\% improved Pile test loss and 1.7\% higher average downstream task accuracy compared to our SP models. Here, we also show that \textmu P performance scales more predictably, enabling more accurate performance extrapolation.

\begin{figure}[h]
    \centering
    \includegraphics[width=0.5\linewidth]{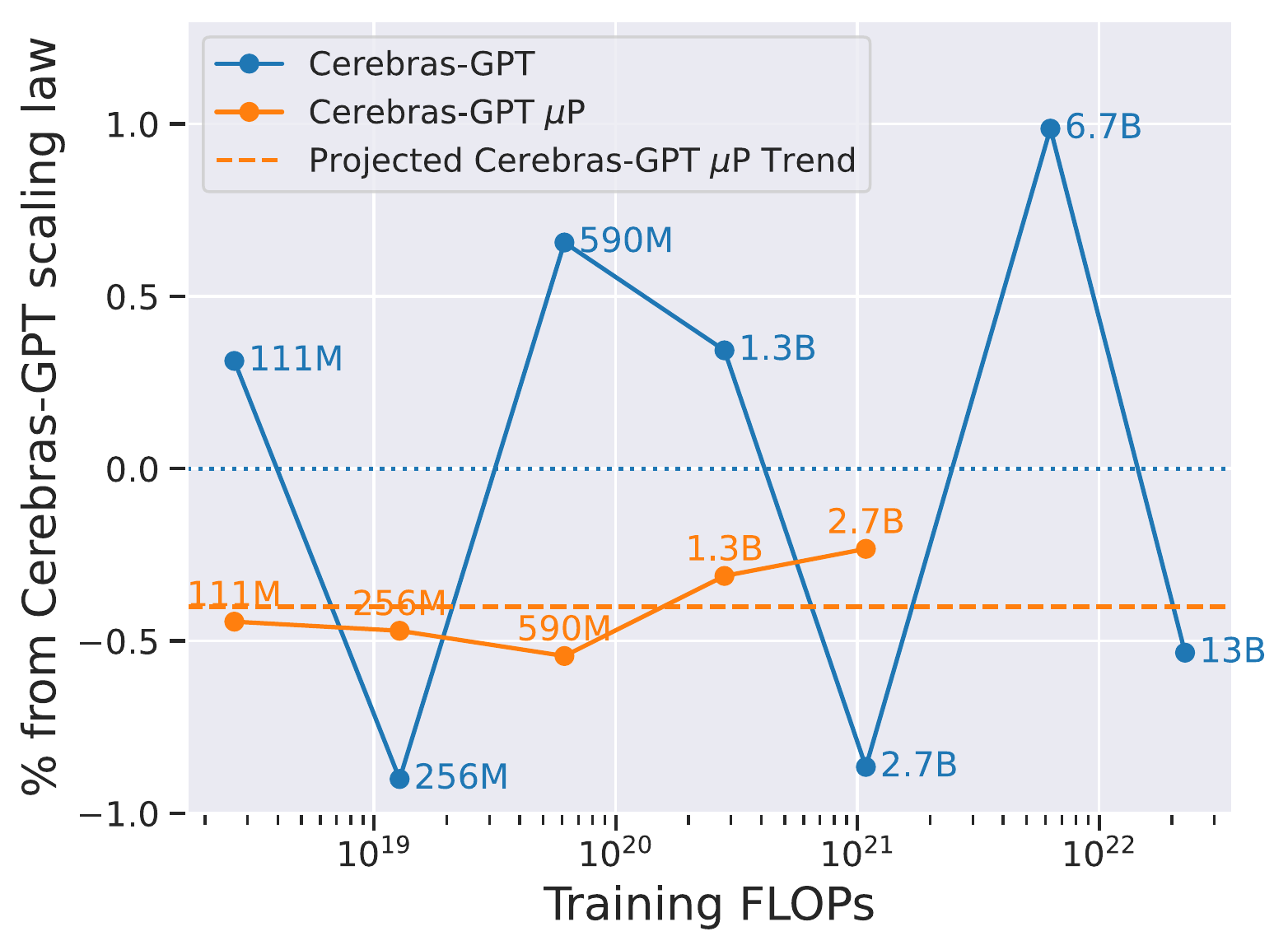}
    \vspace{-6pt}
    \caption{Percentage loss increase relative to Cerebras-GPT scaling law plotted against training FLOPs.}
    \label{figure:mup_improves_scaling_law}
\end{figure}

As we scaled up models with SP, we found model training could become unstable when configured with hyperparameters used in other prior works. At different model size scales, the numerical characteristics of different layers can training instability. These instabilities can lead the practitioner to adjust prior hyperparameters in an effort to work around the issues\footnote{Appendix \ref{appendix:stable_training} describes example stability challenges, such as FP16 mixed precision training causing numerical underflows.}. However, moving away from known good configurations can lead to costly tuning efforts and blocked scaling progress. By shifting to \textmu P, we find more stable training dynamics--key metrics like weight and gradient norms behave similarly at different scales.

We see the benefits of \textmu P readily as we scale. First, after tuning hyperparaeters with a small 40M parameter model, we were able to use the same learning rate hyperparameters for all model scales, as we noted in Table \ref{table:model_details}. \textmu P features were the only changes we made to these models, so scaling was very simple.

Second, models show significantly more predictable scaling. Figure \ref{figure:mup_improves_scaling_law} plots the percentage loss increase for each SP and \textmu P model relative to the SP scaling law (negative values are improved loss). \textmu P models show an average of 0.43\% better Pile test loss compared to the Cerebras-GPT SP scaling law fit. Further, \textmu P models show substantially lower variance with just 0.04\% standard deviation relative to the SP scaling law, while SP models show deviation 0.66\% (${\sim}16\times$ more noisy). For perspective, the run-to-run standard deviation in loss when using different initialization and data random seeds is around 0.35\%.

\begin{table}[h]
\caption{Pile pre-training test loss and zero-shot downstream task results for \textmu P and SP models.}
\vspace{-6pt}
\label{table:mup_sp_eval_results_zero_shot}
\resizebox{\textwidth}{!}
{
\begin{tabular}{lr|C{1.4cm}|C{1.3cm}cC{1.3cm}cccC{1.3cm}C{1.3cm}}
    \thickhline
        &    & Pre-train & \multicolumn{8}{c}{Downstream task accuracy ($\uparrow$)} \\
    \multicolumn{2}{c|}{Model} & Pile test xent ($\downarrow$) & Hella- Swag &     PIQA & Wino- Grande &  Lambada &    ARC-e &    ARC-c & OpenBookQA & Downstream Average \\
    \hline
    Cerebras-GPT             & 111M &          2.608 & \textbf{0.268} &          0.594 &          0.488 &          0.194 &          0.380 &          0.166 &          0.118 &          0.315 \\
    Cerebras-GPT + \textmu P & 111M & \textbf{2.588} & \textbf{0.268} & \textbf{0.598} & \textbf{0.519} & \textbf{0.204} & \textbf{0.390} & \textbf{0.176} & \textbf{0.124} & \textbf{0.325} \\
    \hline
    Cerebras-GPT             & 256M & \textbf{2.349} & \textbf{0.274} &          0.613 & \textbf{0.511} & \textbf{0.293} &          0.410 &          0.170 & \textbf{0.158} &          0.347 \\
    Cerebras-GPT + \textmu P & 256M &          2.359 & \textbf{0.274} & \textbf{0.617} &          0.505 &          0.287 & \textbf{0.427} & \textbf{0.194} &          0.156 & \textbf{0.351} \\
    \hline
    Cerebras-GPT             & 590M &          2.181 &          0.291 &          0.627 &          0.498 & \textbf{0.366} &          0.464 &          0.190 &          0.158 &          0.370 \\
    Cerebras-GPT + \textmu P & 590M & \textbf{2.155} & \textbf{0.295} & \textbf{0.644} & \textbf{0.517} &          0.362 & \textbf{0.470} & \textbf{0.194} & \textbf{0.172} & \textbf{0.379} \\
    \hline
    Cerebras-GPT             & 1.3B &          1.997 &          0.325 &          0.664 & \textbf{0.521} &          0.462 &          0.508 & \textbf{0.224} &          0.166 &          0.410 \\
    Cerebras-GPT + \textmu P & 1.3B & \textbf{1.984} & \textbf{0.334} & \textbf{0.682} &          0.512 & \textbf{0.471} & \textbf{0.515} &          0.223 & \textbf{0.196} & \textbf{0.419} \\
    \hline
    Cerebras-GPT             & 2.7B & \textbf{1.834} &          0.386 & \textbf{0.701} & \textbf{0.559} & \textbf{0.567} & \textbf{0.571} & \textbf{0.246} &          0.206 & \textbf{0.462} \\
    Cerebras-GPT + \textmu P & 2.7B &          1.846 & \textbf{0.388} &          0.697 &          0.557 &          0.558 &          0.569 &          0.241 & \textbf{0.218} &          0.461 \\
    \hline
\end{tabular}
}
\end{table}

In addition to its pre-training advantages, \textmu P also improves downstream capabilities of these models. In the previous Figure \ref{figure:downstream_results}, we plotted downstream results for \textmu P models, where we see improved accuracy and distinctively smoother scaling than SP models. Table \ref{table:mup_sp_eval_results_zero_shot} also lists these zero-shot downstream results for SP and \textmu P models. In particular, \textmu P models show a 1.7\% relative improvement in downstream tasks on average. These results are robust across model scales besides the 2.7B parameter model. We believe that we were just lucky when choosing the SP 2.7B model hyperparameters such that it performs significantly better than the SP Pile scaling law. Despite the SP model's upstream advantage, however, the 2.7B + \textmu P model still performs as well on downstream tasks on average.

\section{Trading Off Training and Inference FLOPs}
\label{section:training-and-inference}

Up to this point, our analysis has focused on compute-optimal pre-training, where compute cost is proportional to the square of the model's size, because we train models to a constant number of tokens per parameter. However, recent work has started to also consider model inference costs, showing smaller models trained on more tokens can still significantly improve loss \citep{hoffmann2022chinchilla,llama}. At inference time, the compute cost is proportional to the model's size and number of inferences. Thus, smaller models will have an overall inference cost advantage proportional to their size.

We propose a technique to identify training+inference compute-optimal frontiers that practitioners can use to estimate how they should pre-train their models when considering inference deployment costs. Specifically, we define a compute cost metric equal to pre-training FLOPs added to the model's inference cost and the expected number of inference tokens. Here, $F$ is the total compute cost, $f$ represents FLOPs costs for full pre-training and per-token inference, $n_{\text{infer\_tokens}}$ is the number of expected inference tokens for the given model, and $p$ is the model's parameter count\footnote{Note that the big-$\mathcal{O}$ order relations here could incorporate constant factors to account for model compression, quantization, or other techniques that decrease the relative inference costs.}:
\begin{equation}
\begin{split}
    F & = f_{\text{pre-train\_total}} + n_{\text{infer\_tokens}} \cdot f_{\text{infer/token}} \\
      & \propto \mathcal{O}(p^2) + n_{\text{infer\_tokens}} \cdot \mathcal{O}(p) \\
\end{split}
\end{equation}

\begin{figure}[h]
    \centering
    \includegraphics[width=1.0\linewidth]{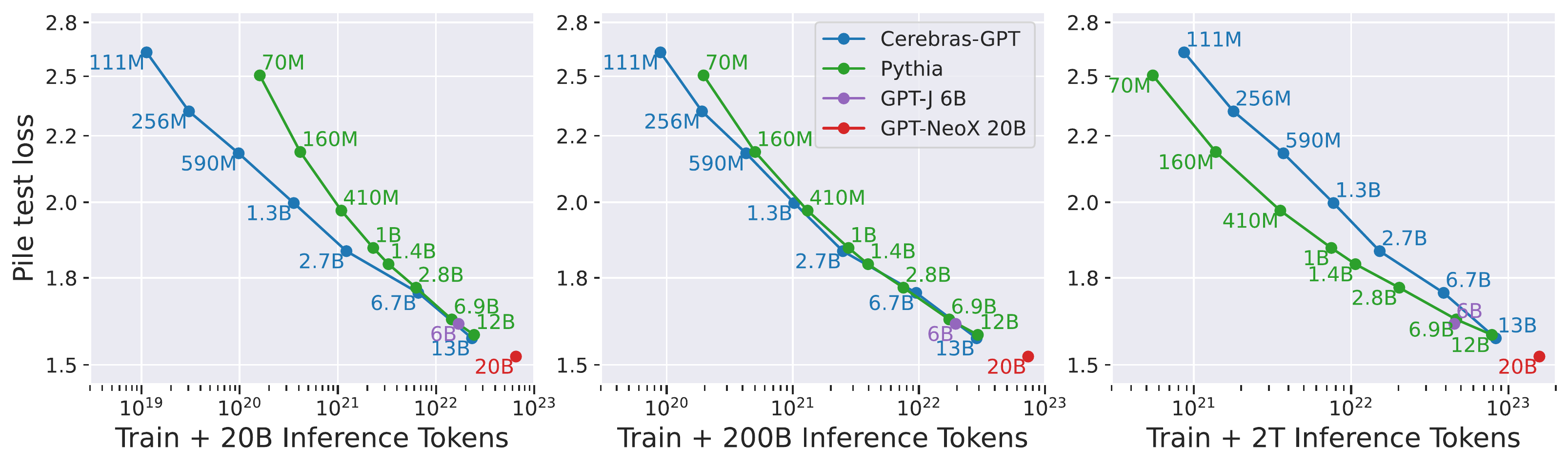}
    \vspace{-6pt}
    \caption{\centering Pile test loss when accounting for both pre-training and expected inference FLOPs. Plots account for 20B (left), 200B (middle), 2T (right) tokens inference.}
    \label{figure:total_compute_costs}
\end{figure}

With this formulation, we can estimate the number of model inferences before the total compute budget matches models trained on fewer or more tokens. Figure \ref{figure:total_compute_costs} plots a comparison of total pre-train + inference compute cost for Cerebras-GPT, GPT-J, GPT-NeoX, and Pythia models assuming either 20B, 200B, or 2T inference tokens. These results show that most Cerebras-GPT models would provide better Pile test-loss-per-compute-FLOP than Pythia models until all models reach roughly 200B inference tokens. Since this total compute metric forms a continuum trade-off, models pre-trained on some number of tokens in between the Cerebras-GPT and Pythia frontiers are likely to achieve better loss for the same total compute budget.

Following this formulation, organizations and governments can better assess the total costs when budgeting large-scale training runs. Specifically, if a model is to be trained in a pre-training compute-{\it inefficient} way using too many data samples, that model may need to be used in a very large number of inferences before the training compute cost can be amortized and well-justified. Similar analysis can be applied to monetary, energy, or carbon footprint costs as well. We encourage the community to consider these total costs when training future models.

\section{Cerebras Stack}

To collect our compute-efficient LLM scaling laws, we run all studies on the Cerebras Wafer-Scale Cluster named ``Andromeda'', which contains 16 Cerebras CS-2 systems. As far as we are aware, this is the first scaling laws study performed on Cerebras systems, which are capable of simple large-scale model training and high-performance scale-out to many systems. In this section, we describe the Andromeda AI Supercomputer, and the Cerebras software platform (CSoft) that we use for scaling and training. We show that Andromeda performance scales linearly up to the full 16 CS-2s, and we describe the simplicity of training models for this study.

\subsection{Andromeda AI Supercomputer}

\begin{figure}[t]
    \vspace{-6pt}
    \centering
    \includegraphics[width=\linewidth]{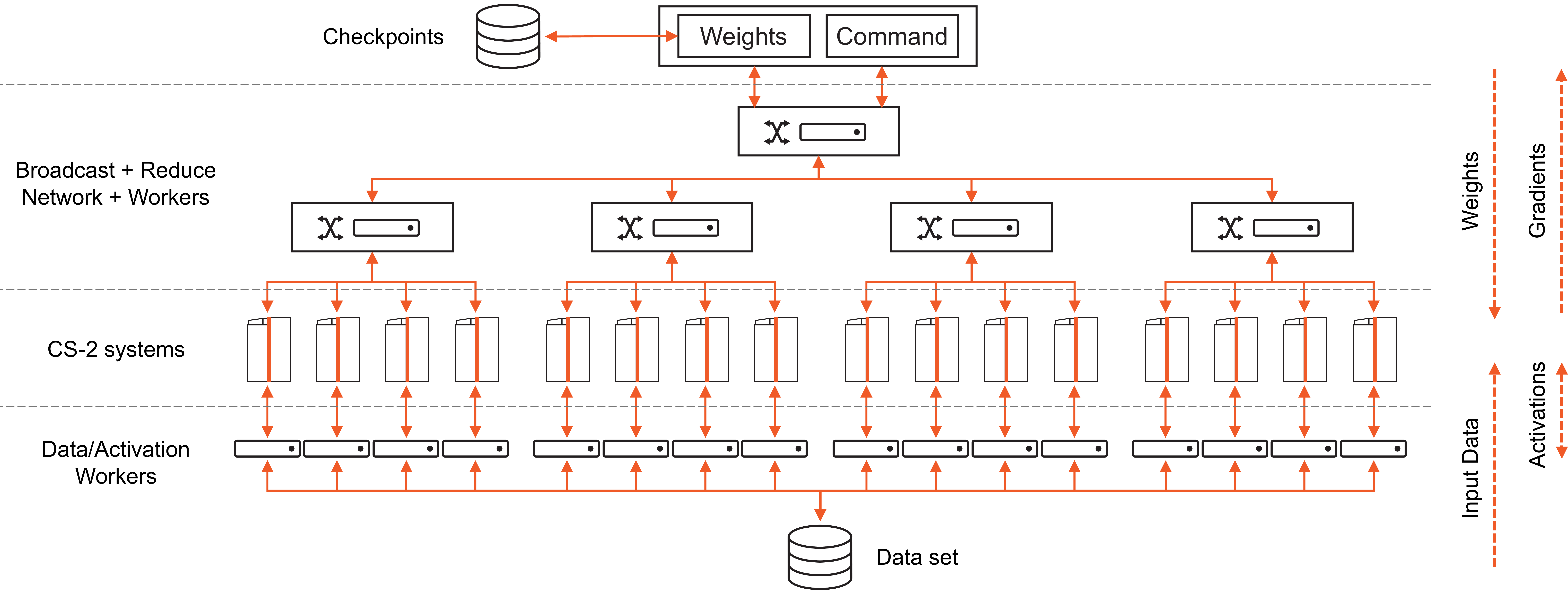}
    \caption{Andromeda AI Supercomputer: logical architecture of the Cerebras Wafer-Scale Cluster.}
    \label{fig:andromeda}
    \vspace{-6pt}
\end{figure}

Andromeda is a Cerebras Wafer-Scale Cluster composed of 16 CS-2 systems. Figure \ref{fig:andromeda} shows the architecture of Andromeda, which aligns well with the large-scale parallel nature of deep learning training. Each CS-2 system contains a Cerebras Wafer-Scale Engine (WSE-2) processor, which has 40 GB of high bandwidth SRAM and compute capability of 7.5 PetaFLOP/s half precision peak throughput. The WSE-2's processing cores are specifically designed to perform all compute operations required for deep learning models. Overall, Andromeda has peak throughput of 120 PFLOP/s from these CS-2s.

Weights and command servers drive the CS-2's computation by broadcasting the weights and control instructions through a broadcast + reduce tree network. This same network collects and reduces gradients from the CS-2s for each training step. When weights servers receive the reduced gradients, they perform the optimizer step and update model weights. They also save and restore model checkpoints to/from disk.

Activation workers act as servers to handle input data and activations. Each worker reads an independent shard of the dataset from disk and creates subbatches to send to a corresponding CS-2 for training. In cases where models must be trained using activation checkpointing, the CS-2 can evict the activations to the corresponding activation worker, which can later refill the activation on the CS-2 when needed.

\subsection{CSoft Platform and Weight Streaming Mode}

Andromeda runs deep learning applications through the Cerebras Software Platform (CSoft). For this study, we write and train models in both Tensorflow and PyTorch (reported results are with PyTorch), and CSoft compiles and orchestrates running these models on the hardware. In this process, CSoft automatically selects things like data parallel subbatch sizing and gradient accumulation, activation recomputation and checkpointing, and appropriate data layouts and kernel configurations for high performance.

The logical data flow in Figure \ref{fig:andromeda} is called the Weight Streaming mode, because weight servers stream the weights to the CS-2s and collect gradients on each training step. This execution mode permits training models of size only limited by the memory capacity of weight servers, and we have tested the ability to train beyond the full GPT-3 175B parameter model with no changes outside of model configurations.

The Weight Streaming design stands in contrast with existing accelerator execution modes. Recent trends in large language model training typically require parallelizing training across tens to thousands of accelerator devices, such as GPUs. These efforts require complicated combinations of data and model parallelism (e.g., \citep{megatron-turing}). Models must be carefully divided to fit into memory close to the devices to achieve high throughput at relatively small per-device batch sizes. Weight Streaming permits moving the weights to the wafer and gradients from the wafer---achieving solid performance at small per-system batch sizes---without the need for model parallelism.

\subsubsection*{Our Language Model Scaling Experience}
We find CSoft Weight Streaming to be significantly easier to develop and scale models than existing accelerator approaches. First, we were able to run each Cerebras-GPT model and even larger models for many training steps on a single CS-2 system. This capability made it easy to quickly test that features of our model and dataset loader implementations would work well even for very large models. Second, the cluster's near-linear performance scaling meant that we could accurately estimate total training time for each run as we scaled to more CS-2 systems. Finally, it was easy to configure these large-scale runs; Scaling to many CS-2 systems requires changing only the number of systems on which to train, and CSoft automatically chooses the data parallel configurations for us.

\subsection{Performance Scalability}
Andromeda provides near-linear performance scaling up to the full 16 CS-2s. We show performance (training speed) scaling from our initial model tests, followed by performance scaling results from our actual training runs. First, as Andromeda came online, we tested performance using a weak scaling approach: As we increased the number of systems, we increase the batch size proportionally (here, batch size is number of sequences of length 2048). We ran 100 training steps for each configuration and take an average training step time over the 100 steps. Table \ref{table:mb_weak_scaling} shows the weak scaling performance relative to 1 CS-2. Andromeda achieves linear scaling within $9\%$ for all model sizes and CS-2 system counts.

\begin{table}[h]
    \centering
    \caption{Andromeda weak scaling tests show linear performance scaling up to 16 CS-2s}
    \vspace{-6pt}
    \label{table:mb_weak_scaling}
    \begin{tabular}{l|rr|rrrr}
                            & Sequence & Per CS-2   & \multicolumn{4}{c}{Performance relative to 1 CS-2} \\
        Model               & Length   & Batch Size & 2 CS-2s & 4 CS-2s & 8 CS-2s & 16 CS-2s \\
        \hline
        GPT-3 XL 1.3B       &    2,048 & 121        &   1.99x &   3.94x &   7.87x & 15.50x \\
        GPT-3 XL 1.3B       &   10,000 & 33         &   1.99x &   3.97x &   7.95x & 15.87x \\
        GPT-3 2.7B          &    2,048 & 121        &   1.98x &   3.91x &   7.86x & 15.62x \\
        GPT-3 6.7B          &    2,048 & 85         &   1.99x &   3.89x &   7.91x & 15.45x \\
        GPT-3 20B           &    2,048 & 50         &   1.92x &   3.75x &   7.93x & 15.32x \\
        GPT-J 6B            &    2,048 & 65         &   1.97x &   3.65x &   7.69x & 14.52x \\
        GPT-NeoX 20B        &    2,048 & 50         &   1.98x &   3.92x &   8.05x & 15.45x \\
    \end{tabular}
\end{table}

We also show that Andromeda achieves high utilization even when strong scaling on batch size. We choose to scale out the fixed batch sizes from our training runs across different numbers of Andromeda systems. When running on fewer CS-2s, if the per-CS-2 batch size requires too much memory to fit in each WSE-2's on-wafer SRAM, the software stack automatically selects a smaller per-CS-2 batch size and accumulates gradients up to the user's chosen batch size. Table \ref{table:mb_strong_scaling} lists the relative performance compared to running on a single CS-2. These results show consistent performance scalability for the batch sizes commonly chosen for these models.

\begin{table}[h]
    \centering
    \caption{\centering Strong scaling performance for batch sizes used to train larger models. To get to the user's full batch size, CSoft uses data parallelism across systems and gradient accumulation.}
    \label{table:mb_strong_scaling}
    \vspace{-6pt}
    \begin{tabular}{l|c|cccc}
                &            & \multicolumn{4}{c}{Performance relative to 1 CS-2 (per CS-2 batch)} \\
        Model   & Batch Size & 1 CS-2 & 2 CS-2s & 4 CS-2s & 8 CS-2s  \\
        \hline
        1.3B    &        528 &   1.0x (132) & 1.99x (132) &   3.97x (132) & 7.10x (66) \\
        2.7B    &        528 &   1.0x (88)  & 1.99x (88)  &   3.77x (66)  & 7.43x (66) \\
        6.7B    &      1,040 &   1.0x (65)  & 1.99x (65)  &   3.97x (65)  & 7.90x (65) \\
        13B     &      1,040 &   1.0x (65)  & 1.99x (65)  &   3.95x (65)  & 7.84x (65) \\
    \end{tabular}
\end{table}

Finally, as we increased model sizes along our scaling law, we tested and compared the cluster's FLOP/s utilization for each training run. Table \ref{table:model_size_scaling_perf} lists Andromeda's utilization relative to the 111M parameter model running on one CS-2. Performance deviates by less than 8\% at all model scales. In addition to robust scaling across many machines, these results indicate consistent performance across a range of model and batch sizes.

\begin{table}[h]
    \centering
    \caption{\centering Andromeda FLOP/s utilization relative to 1 CS-2 training the 111M parameter model. Here, larger values mean higher utilization.}
    \vspace{-6pt}
    \label{table:model_size_scaling_perf}
    \begin{tabular}{l|cccc}
              &            & Number   & Per CS-2   & Relative \\
        Model & Batch Size & of CS-2s & Batch Size & Utilization \\
        \hline
        111M  &        120 &        1 &        120 & 1.00 \\ %
        256M  &        264 &        1 &        264 & 1.00 \\ %
        590M  &        264 &        1 &        264 & 0.92 \\ %
        1.3B  &        528 &        4 &        132 & 0.96 \\ %
        2.7B  &        528 &        4 &        132 & 0.96 \\ %
        6.7B  &       1040 &       16 &         65 & 1.05 \\ %
        13B   &       1080 &       12 &         45 & 1.02 \\ %
    \end{tabular}
\end{table}

\section{Related Work}

Early deep learning scaling law studies show that when scaling dataset and model size, loss improves predictably \citep{hestness2017scalinglaws,kaplan2020scalinglaws}. These studies indicate generally that scaling could give substantial modeling improvements. From this observation, many organizations scaled to train the largest possible models on their available infrastructure:
  GPT-3 175B \citep{gpt3},
  Jurassic-1 178B \citep{jurassic},
  Gopher 280B \citep{gopher},
  HyperCLOVA 82B \citep{hyperclova},
  Ernie 3.0 Titan 260B \citep{ernie3titan},
  Yuan 1.0 \citep{Wu2021Yuan1L},
  PanGu-$\alpha$ \citep{pangu-alpha},
  Megatron-Turing NLG 530B \citep{megatron-turing},
  PaLM 540B \citep{Chowdhery2022PaLMSL},
  and
  LaMDA 137B \citep{lamda}.
These models show significant performance improvement on many downstream tasks compared to prior language models. However, these studies typically only scale model size without scaling the dataset size as suggested by the early works, often training on roughly 300B tokens. Further, these models could only be trained by select organizations with large compute clusters, and the datasets and resulting pre-trained models have not been released publicly for analysis by the research community.

The research community \textit{has} released large datasets and pre-trained models---typically much smaller than the largest models above but still quite valuable---and we have noted many of them previously: GPT-J, GPT-Neo, GPT-NeoX, OPT, and Pythia. Another notable work that releases dataset and model is the Big Science collaborative effort to train BLOOM 176B \citep{1m-gpu-hours,bloom176b}. These studies, datasets, and models enable the community to test, compare, and use large language models they would otherwise not have access to or compute budget to train.

In 2022, studies started revisiting early scaling works to note that although model size scaling improves performance, consistently scaling the dataset size is still critical to get the best possible models. \cite{hoffmann2022chinchilla} show that for compute-optimal pre-training, the dataset size should grow linearly with transformer model size in parameters, and they scaled training up to a 70B parameter model on 1.4T tokens. The dataset and models are not publicly-available, so our work aims to reproduce these results to offer to the community an open and reproducible scaling law. Recently, the LLaMa paper \citep{llama} also reproduces large models trained on large open datasets to improve pre-training. Although these models show strong performance, most are trained in a compute-inefficient way by training on larger datasets than would be compute-optimal for the given model sizes. LLaMa models are available through request. The resulting models trained in these works perform better than prior larger models trained on smaller datasets.

Large language model training is prone to instability, and it is very costly when large model training runs fail due to instability. Various techniques have been developed to control training dynamics and train models stably  \citep{glorot2010initialization,yang2017chaos,schoenholz2017deep,yang2018deep,zhang2018residual,bachlechner2020rezero,pmlr-v119-huang20f,liu-etal-2020-understanding,li2022the}. \textmu P is the first comprehensive method to analytically control width-related training instabilities and allow optimal hyperparameters of small models to be the same as optimal hyperparameters for very large models. We find that the comprehensive nature of \textmu P simplifies our training efforts, so we feel it is useful to share our experience and encourage the community to use it rather than considering combinations of other techniques.

\section{Limitations}
In this work, we train well-established model architectures to create foundation models, but we did not explore recent architectural features, downstream task tuning procedures, or dataset cleaning approaches used in contemporary works. Model features worth exploring in future work include position embeddings, such as RoPE \citep{su2022roformer} and ALiBi \citep{alibi}, and activation functions, like SwiGLU \citep{shazeer2020glu}. There are also training paradigms worth exploring, such as denoising pre-training objectives \citep{tay2023ul2} and instruction fine-tuning \citep{ouyang2022training}. Finally, we expect that further dataset cleaning can further improve pre-trained models. For instance, our testing in Appendix \ref{appendix:downstream_tasks} shows that the Pythia models improve downstream task accuracy when trained on a deduplicated version of the Pile.

We have not yet tested Cerebras-GPT models extensively in downstream tasks or in real application settings. Specifically, we have not tested for factual accuracy, profanity, toxicity, or other socially undesirable text generation. We do evaluate the bias of our Cerebras-GPT models using the CrowS-Pairs dataset in Appendix \ref{section:bias}. Further safety-related testing, mitigations, and output curation should be applied to our pre-trained models before presenting results to users. Please refer to the model card in the Appendix, Table \ref{table:model_card}.

\section{Conclusion}

In this paper, we introduce Cerebras-GPT, a family of open models scaled from 111M to 13B parameters and pre-trained in a compute-optimal way on the Pile dataset. These models show state-of-the-art pre-training efficiency on pre-training and downstream objectives when compared to other open-source models. We believe this is the first such open effort, and we provide detailed instructions to reproduce our results and we release our pre-trained model checkpoints\footnote{Pre-trained models are available on HuggingFace: \url{https://huggingface.co/cerebras}. Source code is available in the Cerebras Modelzoo: \url{https://github.com/Cerebras/modelzoo}.}. We combine this scaling with \textmu P, a comprehensive technique to improve large model stability, and we show it further improves our scaling results. We document our experience training these models on the Andromeda AI Cluster, comprising 16 Cerebras CS-2 systems, and we describe the simplicity of scaling models and performance.

\bibliography{main}
\bibliographystyle{tmlr}

\newpage
\appendix

{\Large\bf Appendix}
\subsection*{Model Card}
Table \ref{table:model_card} shows the model card for the largest Cerebras-GPT model following guide in \citep{mitchell2018modelcards}.

\begin{table}[h]
\vspace{-10pt}
\centering
{\scriptsize
\caption{Cerebras-GPT 13B Parameter Model Card}
\vspace{-8pt}
\label{table:model_card}
\resizebox{\textwidth}{!}
{
\centering
\renewcommand{\arraystretch}{1.15}
\begin{tabular}{| p{15.8cm} |}
    \hline
    \hline
    \textbf{Release details} \\
    \begin{itemize}
        \vspace{-12pt}
        \setlength\itemsep{-2px}
        \item {\bf Organization}: Cerebras Systems
        \item {\bf Model date}: March 2023
        \item {\bf Model type}: Autoregressive Transformer Language Model (more details in Section \ref{section:training-setup})
        \item {\bf Feedback on the model}: Nolan Dey and Joel Hestness, \{nolan, joel\}@cerebras.net
        \vspace{-4pt}
    \end{itemize} \\

    \hline
    \textbf{Model details} \\
    \begin{itemize}
        \vspace{-12pt}
        \setlength\itemsep{-2px}
        \item {\bf Model architecture}: Cerebras-GPT 13B is an autoregressive transformer decoder-only model with 13 billion parameters. The architecture is similar to GPT-2 and GPT-3. More details in Section \ref{section:architecture}.
        \item {\bf Hidden size}: 5,120
        \item {\bf Number of layers}: 40
        \item {\bf Head size}: 128
        \item {\bf Filter size}: 20,480
        \item {\bf Context (sequence) length}: 2,048
        \item {\bf Initialization}: Model is trained from randomly initialized weights. The base variant uses standard parameterization initialization (see Section \ref{section:mup_method}).
        \item {\bf Release license}: Apache 2.0
        \vspace{-4pt}
    \end{itemize} \\
    \hline

    \textbf{Data Overview} \\
    \begin{itemize}
        \vspace{-12pt}
        \setlength\itemsep{-2px}
        \item {\bf Training data}: Cerebras-GPT is trained on the Pile dataset \citep{pile}
        \item {\bf Pre-processing}: Pile was cleaned using ftfy library to normalize text, and then filtered using scripts provided by Eleuther. Then, data was tokenized with byte-pair encoding using the GPT-2 vocabulary.
        \item {\bf Evaluation data}: Upstream (pre-training) evaluations were completed using the Pile validation and test set splits. Downstream evaluations were performed on standardized tests. Cloze and completion tasks: LAMBADA, HellaSwag. Common Sense Reasoning tasks: PIQA, ARC, OpenBookQA. Winograd schema type tasks: Wino-grande. Downstream evaluations were performed using the Eleuther lm-eval-harness \citep{eval-harness}.
        \item {\bf Motivation}: Evaluation tasks were chosen to closely match related works and cover a broad cross-section of task types.
        \vspace{-12pt}
    \end{itemize} \\
    \hline

    \hline
    \textbf{Usage} \\
    \begin{itemize}
        \vspace{-12pt}
        \setlength\itemsep{-2px}
        \item {\bf Primary intended uses}: The primary intended use is to further research into large language models. Model can be used as a foundation model for NLP, applications, ethics, and alignment research.
        \item {\bf Primary intended users}: Researchers who are working to improve LLMs and practitioners who are looking for reference implementations, training setups, hyperparameters, or pre-trained models.
        \item {\bf Limitations}: Due to financial and compute budgets, Cerebras-GPT models were only trained and evaluated following the approaches described in this document.
        \item {\bf Out-of-scope uses}: Further safety-related testing and mitigations should be applied before using the Cerebras-GPT model family in production downstream applications.
        \vspace{-4pt}
    \end{itemize} \\

    \hline
    \textbf{Metrics} \\
    \begin{itemize}
        \vspace{-12pt}
        \setlength\itemsep{-2px}
        \item {\bf Model performance measures}: Model is evaluated using text prediction cross-entropy on upstream tasks and text generation accuracy on downstream tasks. Results are compared against many publicly available large language models. Details can be found in Section \ref{section-results}.
        \item {\bf Uncertainty and variability}: Model is not evaluated for prediction uncertainty or calibration. Due to restricted compute budget, variability analysis was only performed for small variants of Cerebras-GPT models using multiple runs from different random initializations and data loader seeds to assess variance in task performance.
        \vspace{-4pt}
    \end{itemize} \\

    \hline
    \textbf{Ethical considerations} \\
    \begin{itemize}
        \vspace{-12pt}
        \setlength\itemsep{-2px}
        \item {\bf Data}: The Pile dataset has been thoroughly analyzed from various ethical standpoints, and the dataset is known to contain content considered toxic, gender biased, pejorative, racially sensitive, etc. Please refer to Pile dataset references.
        \item {\bf Human life}: The outputs from this model may or may not align with human values. The risk needs to be thoroughly investigated before deploying this model in a production environment where it can directly impact human life.
        \item {\bf Risks and harms}: There can be distributional bias in the Pile dataset that can manifest in various forms in the downstream model deployment. There are other risks associated with large language models such as amplifying social stereotypes, memorizing training data, or revealing private or secure information.
        \item {\bf Mitigations}: Only mitigations in standard Pile dataset pre-processing were employed when pre-training Cerebras-GPT.
        \vspace{-4pt}
    \end{itemize} \\

    \hline
    \textbf{Factors} \\
    \begin{itemize}
        \vspace{-12pt}
        \setlength\itemsep{-2px}
        \item {\bf Evaluation factors}: Cerebras-GPT was evaluated for various bias factors using the CrowS-Pairs dataset task. Details are in Appendix \ref{section:bias}.
        \vspace{-4pt}
    \end{itemize} \\

    \hline
    \textbf{Implementation infrastructure} \\
    \begin{itemize}
        \vspace{-12pt}
        \setlength\itemsep{-2px}
        \item {\bf Hardware}: Andromeda AI Supercomputer: Cerebras Wafer-Scale Cluster with 16 Cerebras CS-2 systems
        \item {\bf Software}: PyTorch, Cerebras Software Platform (CSoft) release 1.8
        \vspace{-4pt}
    \end{itemize} \\
    \hline
\end{tabular}
} %
} %
\end{table}

\subsection*{Cerebras-GPT Open-Source References}
We release our pre-trained models and code, so the community can use and reproduce our results. Pre-trained models are available on HuggingFace: \url{https://huggingface.co/cerebras}. We are initially releasing seven Cerebras-GPT models with 111M, 256M, 590M, 1.3B, 2.7B, 6.7B, and 13B parameters trained with standard parameterization (SP). These models are released under Apache 2.0 license, which permits commercial and non-commercial use. Source code is available in the Cerebras Modelzoo: \url{https://github.com/Cerebras/modelzoo}. We hope these models will be a valuable addition for the open-source community.

\subsection*{Author Contributions and Acknowledgements}

We would like to acknowledge the contributions of those who helped in preparation of this manuscript.

\textbf{Experimental planning and strategy:} Nolan Dey, Joel Hestness \\
\textbf{Model training:} Zhiming (Charles) Chen, Hemant Khachane, Ribhu Pathria, Gurpreet Gosal \\
\textbf{Dataloader development and dataset preparation:} Gurpreet Gosal \\
\textbf{Numerical configuration and validation:} Joel Hestness, Hemant Khachane, Gurpreet Gosal \\
\textbf{Upstream loss comparisons:} Gurpreet Gosal, Charles Chen \\
\textbf{Downstream task comparisons:} William Marshall \\
\textbf{Manuscript preparation:} Nolan Dey, Joel Hestness, Gurpreet Gosal, William Marshall \\
\textbf{Overall project leadership:} Joel Hestness, Marvin Tom \\
\textbf{Overall technical leadership:} Joel Hestness

In addition, we would like to thank others who helped in the preparation of this work. Bowen Yang and Faisal Al-Khateeb helped prepare the Pile dataset. We are also thankful for helpful feedback on the manuscript provided by Sean Lie, Anshul Samar, and Vithu Thangarasa. In general, we would also like to acknowledge the contributions of the many Cerebras engineers who made this work possible.

\section{Methods Details}

\subsection{Pile Dataset Preprocessing \label{appendix_dataset}}

We preprocess Pile using tools and instructions provided by Eleuther and the community. We clean the raw text data sources using the ftfy library to normalize text, including cleaning corrupted unicode \citep{speer-2019-ftfy}. Our tokenized version of the Pile training set contains roughly 371B tokens (validation 380M, test 371M), similar to results reported in the GPT-NeoX paper \citep{black2022gptneox}. The resulting tokenized dataset files contain contiguous samples from the raw text. For the best model generalization, we find it critical to shuffle samples across all training set documents, rather than shuffling within a window of even a few thousand documents. So, we also shuffle the training dataset across all documents as a final preprocessing step. This dataset-wide shuffling improves validation loss by 0.7-1.5\% compared to aggressive shuffling settings over sets of contiguous documents as we tested with our dataloaders.

The Pile dataset has been thoroughly analyzed from various ethical standpoints, and the dataset is known to contain content considered toxic, gender biased, pejorative, racially sensitive, etc. Please refer to Pile dataset references for further information.

\subsection{Ensuring Stable Training}
\label{appendix:stable_training}

As we scaled up models to larger sizes, we encountered and resolved a few issues that improve training stability. We share some details here in hopes they assist others in their scaling efforts.

\textbf{Mixed Precision Training:} Initially, we trained models using FP16 mixed precision, a technique that carries model weights and activations in IEEE half precision floating-point (FP16) while performing dot-products and reductions in single precision 32-bit (FP32). This approach ensures that reductions maintain precision, while taking advantage of the smaller 16-bit data format for storing activations. Because FP16 has a significantly reduced exponent range compared to FP32, models need to be trained with loss scaling, an approach that multiplies the gradients by a large positive value before back-propagation, and then divides out this multiplier just before applying the calculated gradients to the weights in the optimizer step. A dynamic approach to loss scaling sets the scale value by periodically testing larger values to see the largest scale such that the gradients do not overflow FP16.

We found that for models larger than roughly 1.3B parameters (hidden size 2048), loss scaling alone was not sufficient to ensure stable model training. As model weights grow during training, the gradients through layers like softmaxes can become eccentric, leading to large gradient values that tend to overflow. These large values push down the maximum allowed loss scale and cause other gradients to be very small. Very small gradient values have a tendency to underflow in FP16. Underflow can cause weights to receive either no gradient or low-precision, eccentric gradients, which can further exacerbate dynamic loss scale and underflow.

\textbf{Underflows and Weight Growth:} We detect underflows by observing any significant increase in the number of identically zero values in tensors as they go through cast operations from FP32 to FP16. Specifically, we find that attention layer softmax gradients are particularly susceptible to underflow. To fix this issue, we recommend carrying gradients in FP32 from the softmax back through the corresponding query-key dot-product and when calculating the gradients for the query and key projection weights and biases. We have tested various open-source mixed precision attention implementations that suffer this same issue.

We also find specific layers to be most susceptible to eccentric gradients caused by underflow. In the attention layers, the bias weights of the keys projection, specifically, have expected value close to zero early in training. If gradients to these weights partially underflow, the remaining gradients will be eccentric and large relative to their expectation. These K bias weights will tend to grow very quickly under these circumstances. We detect this issue by inspecting weight growth---measuring the weight standard deviation and norms---over many training steps compared to an implementation that uses FP32.

\textbf{Switching to bfloat16:} Another approach to avoid underflows is to use a larger exponent range for activation and gradient tensors. Brain floating-point (bfloat16) is a numerical format introduced by Google Brain and used in various hardware platforms to improve half precision floating point range. Specifically, bfloat16 has 8 bits of exponent compared to 5 bits for FP16. Typical bfloat16 model training implementations still use FP32 for intermediate values (mixed precision) in reduction operations to ensure mantissa precision.

Bfloat16 eliminates the need for dynamic loss scaling that is used with mixed precision, because the exponent range significantly reduces the likelihood of underflows. We find that although bfloat16 does not completely eliminate low-precision training dynamics concerns, it does significantly improve training stability, so we use bfloat16 for all final models that we train in this paper and release publicly. We find that our experience with bfloat16 training stability is consistent with prior works.

\textbf{Setting Adam Epsilon:} When gradients for a set of weights are small, using a relatively large Adam epsilon value can cause weights to grow slowly. This might be an appealing approach in the presence of large weight growth among weights that are expected to be small. However, a large Adam epsilon can cause very poor weights resolution and degrade model quality. Given the Adam update at step $t$ on weights $\theta$:

\vspace{-18pt}
\begin{equation}
    \vspace{0pt}
    \begin{split}
        \theta_t & = \theta_{t-1} - \gamma m_t / (\sqrt{v_t} + \epsilon) \\
    \end{split}
\end{equation}
\vspace{-18pt}

Here, $m_t$ is the momentum, a running average of the gradient, and $v_t$ is the velocity, a running average of the squared gradient. When gradients to a weight are small (e.g., in the case of K bias weights growth above), $v_t$ will tend to be very small, because it is squared. In this case, $\epsilon$ needs to be chosen to be small relative to each $\sqrt{v_t}$, or the Adam update denominator will be large, causing the weight updates to be small. As a rule-of-thumb, we find $\epsilon$ should be less than $\sqrt{\mu_v}/1000$, where $\mu_v$ is the mean of the velocity state weights, to ensure models do not suffer from stagnant weight growth. This analysis is how we choose to lower epsilon for our 6.7B and 13B parameter models.

\section{Downstream Task Details}
\label{section:downstream_tasks}
We evaluate our models on the following six downstream tasks in both the zero-shot and the few-shot setting. Here, we briefly describe each of the tasks: HellaSwag, PIQA, WinoGrande, Lambada, ARC, and OpenBookQA.
\lstset{
    basicstyle=\small\ttfamily,
    columns=flexible,
    breaklines=true
}
\begin{enumerate}
    \item \textbf{HellaSwag} is a dataset of multiple choice questions aimed to test a model's common sense reasoning abilities \citep{hellaswag}. For example,
    \begin{lstlisting}
    A woman is outside with a bucket and a dog. The dog is running around trying to avoid a bath. She...
    A. rinses the bucket off with soap and blow dry the dog's head.
    B. uses a hose to keep it from getting soapy.
    C. gets the dog wet, then it runs away again.
    D. gets into a bath tub with the dog.
    \end{lstlisting}
    The authors of the dataset adversely select examples such that they are difficult for language models while still trivial for humans (with reported greater than 95\% accuracy).

    \item \textbf{PIQA} tests a model's common sense reasoning about the physical world by posing a prompt and two potential completions \citep{piqa}. For example
    \begin{lstlisting}
    [Goal] Make an outdoor pillow
    [Sol1] Blow into a tin can and tie with rubber band
    [Sol2] Blow into a trash bag and tie with rubber band
    \end{lstlisting}
    The model must choose which of the two continuations is more likely to follow from the prompt. Human performance on this dataset is approximately 95\%.

    \item \textbf{WinoGrande} consists of a set of pronoun resolution problems \citep{winogrande}. Samples are constructed as pairs of similar sentences, each with a pronoun referring to a noun earlier in the sentence. The task is to predict which noun the pronoun refers to. For example, in the sample
    \begin{lstlisting}
    a. The trophy doesn't fit into the brown suitcase because it's too large.
    b. The trophy doesn't fit into the brown suitcase because it's too small.
    \end{lstlisting}
    in sentence (a), ``it's'' refers to ``trophy'', while in sentence (b), changing a single context word modifies the meaning of the sentence such that ``it's'' now refers to ``suitcase''.

    \item \textbf{Lambada} is a word prediction task that tests a model's ability to understand text, with a particular emphasis on global context \citep{lambada}. For example
    \begin{lstlisting}
    Context: They tuned, discussed for a moment, then struck up a lively jig. Everyone joined in, turning the courtyard into an even more chaotic scene, people now dancing in circles, swinging and spinning in circles, everyone making up their own dance steps. I felt my feet tapping, my body wanting to move.
    Target sentence: Aside from writing, I've always loved ___.
    Target word: dancing
    \end{lstlisting}
    There are two versions of the Lambada dataset. The original version is that which was published in \citep{lambada}. However, researchers more commonly use version of the dataset with slightly different formatting that was created by Radford et al.\ in order to evaluate their GPT-2 model \citep{radford2019gpt2}. In our evaluations we use the latter version, referred to as ``lambada\_openai'' in the Eleuther eval harness \citep{eval-harness}.

    \item \textbf{ARC} tests a model's ability to answer multiple choice science questions \citep{arc}. For example
    \begin{lstlisting}
    Which property of a mineral can be determined just by looking at it?
    (A) luster [correct] (B) mass (C) weight (D) hardness
    \end{lstlisting}
    This dataset is split into an ``easy'' set and a ``challenge'' set where samples are selected for the challenge set if they are answered incorrectly by word co-occurrence and retrieval based algorithms.

    \item \textbf{OpenBookQA} is a multiple choice common sense question answering dataset \citep{openbookqa}. One example question from this dataset is
    \begin{lstlisting}
    What is the most likely to be an effect of acid rain on an aquatic environment?
    (A) increase in plant growth
    (B) increase in fish population
    (C) decrease in plant life
    (D) cleaner and clearer water
    \end{lstlisting}
\end{enumerate}

\section{Additional Results}
\label{section:additional_results}

\subsection{Pre-training Losses Throughout Training}
\begin{figure}[h]
    \centering
    \includegraphics[width=0.7\linewidth]{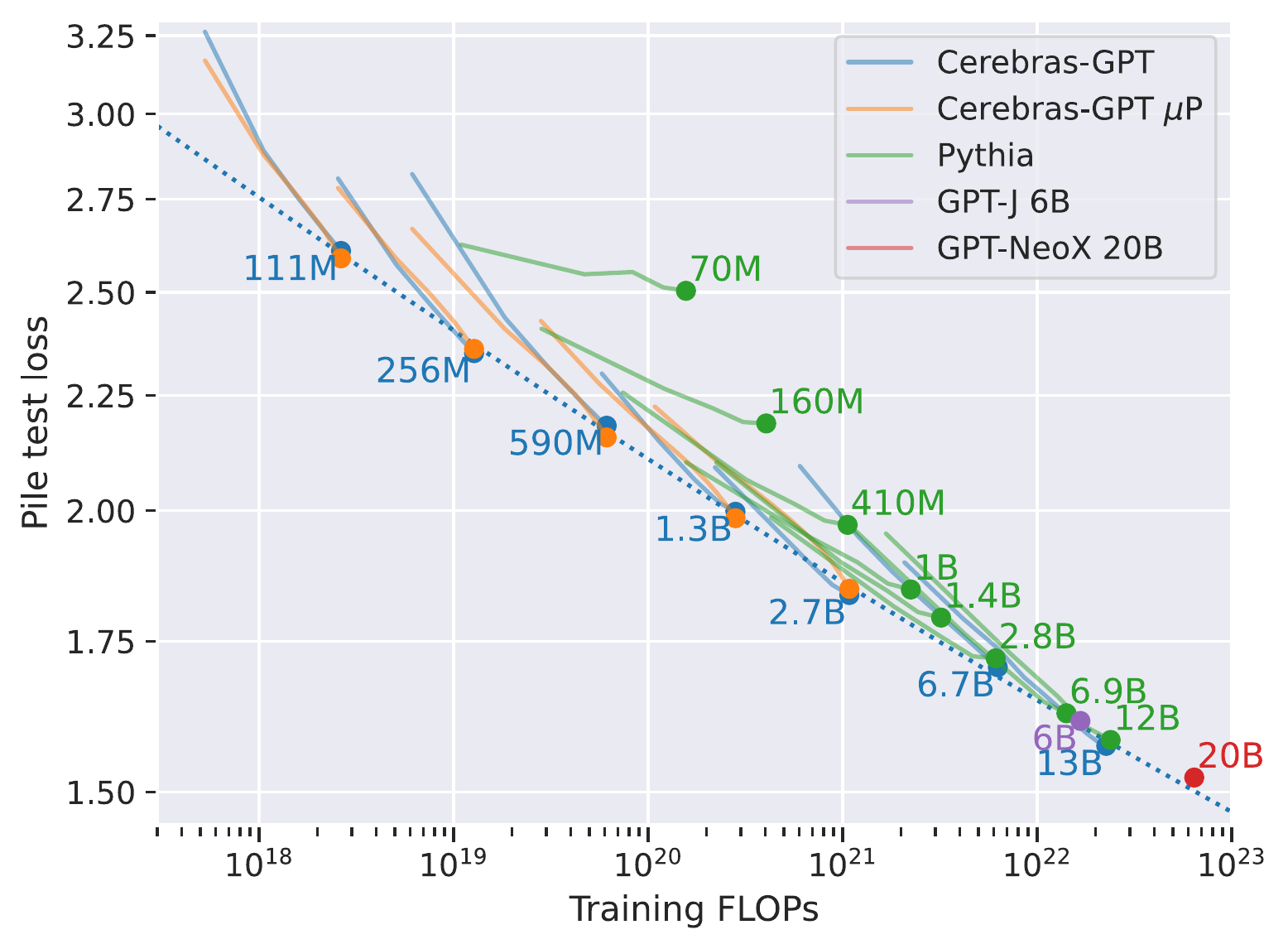}
    \vspace{-6pt}
    \caption{Pile test set loss given pre-training FLOPs throughout training for Cerebras-GPT and Pythia.}
    \label{figure:scaling_laws_intermediate}
\end{figure}

In Figure \ref{figure:scaling_laws_intermediate}, we show the intermediate Pile test losses achieved throughout training for Pythia and Cerebras-GPT models. For all model sizes and compute budgets, Cerebras-GPT models tend to have similar trajectory when approaching their final results along the scaling law. In contrast, Pythia models trained for more tokens per parameter follow less efficient trajectory, trending away from the scaling law and indicating their over-training. For Pythia models trained closer to 20 tokens per parameter, the trajectories align more closely with Cerebras-GPT models.

\subsection{Complete Downstream Task Testing}
\label{appendix:downstream_tasks}

\begin{figure}[p]
    \centering
    \begin{minipage}{0.9\linewidth}
        \centering
        \includegraphics[width=0.9\linewidth]{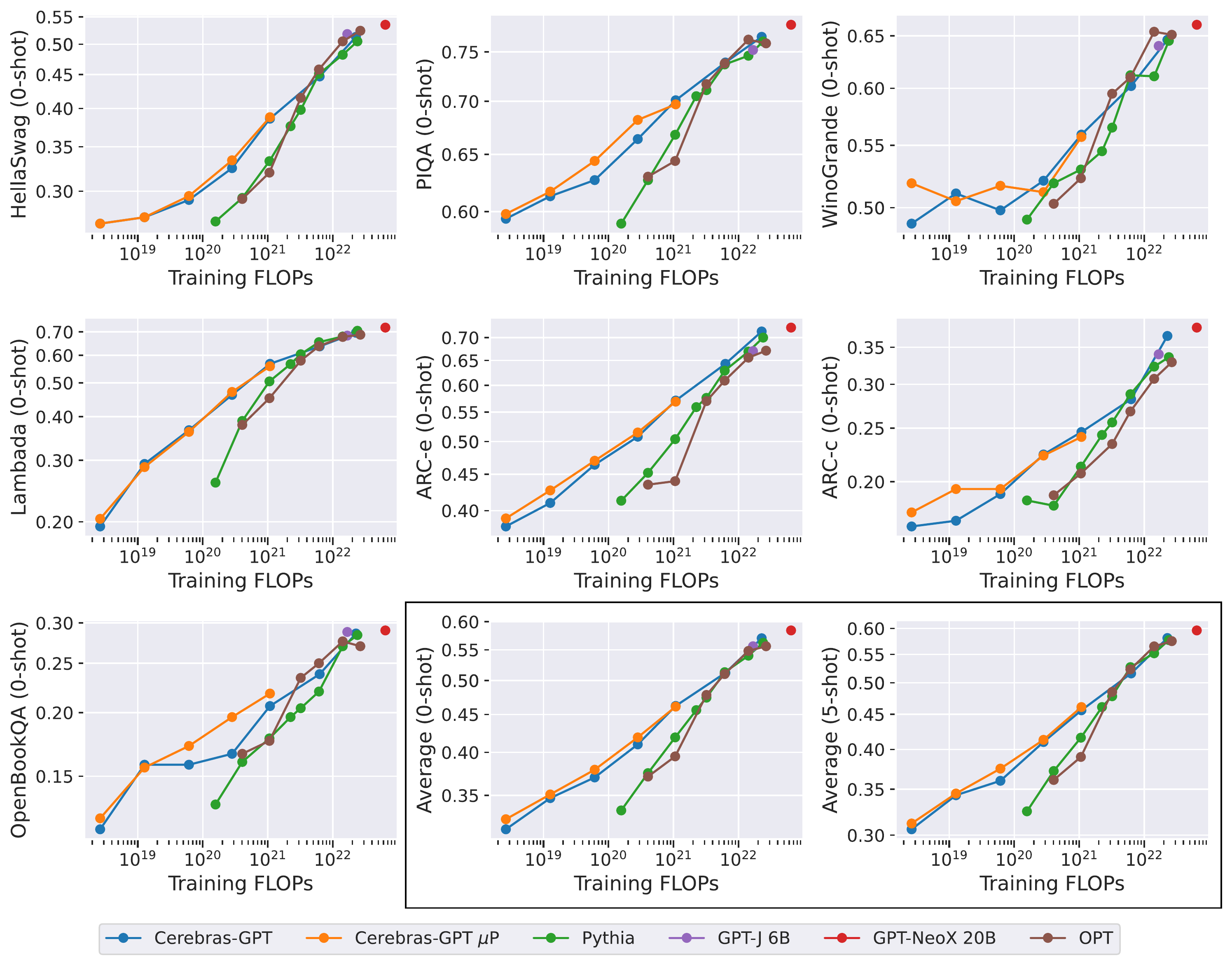}
        \vspace{-6pt}
        \caption{\centering Individual zero-shot downstream task accuracy, zero-shot average, and five-shot average, plotted against training FLOPs.}
        \label{figure:downstream_all_flops}
    \end{minipage} \\
    \vspace{18pt}
    \begin{minipage}{0.9\linewidth}
        \centering
        \caption{Zero-shot downstream task accuracy using length-normalized NLL selection criteria.}
        \vspace{-6pt}
        \label{table:0_shot_acc_norm}
        \begin{tabular}{lr|ccccc}\thickhline
        \multicolumn{2}{c|}{Model} & HellaSwag &     PIQA &    ARC-e &    ARC-c & OpenBookQA\\\hline
        \multirow{7}{*}{Cerebras-GPT} &     111M &    0.271 &    0.581 &    0.348 &    0.206 &    0.278\\
                 &     256M &    0.286 &    0.614 &    0.376 &    0.218 &    0.254\\
                 &     590M &    0.324 &    0.629 &    0.412 &    0.235 &    0.280\\
                 &     1.3B &    0.384 &    0.666 &    0.458 &    0.250 &    0.290\\
                 &     2.7B &    0.488 &    0.707 &    0.525 &    0.273 &    0.318\\
                 &     6.7B &    0.589 &    0.740 &    0.579 &    0.312 &    0.366\\
                 &      13B &    0.684 &    0.776 &    0.673 &    0.395 &    0.406\\\hline
        \multirow{5}{*}{Cerebras-GPT + \textmu P} &     111M &    0.276 &    0.598 &    0.344 &    0.223 &    0.260\\
                 &     256M &    0.287 &    0.618 &    0.376 &    0.225 &    0.258\\
                 &     590M &    0.333 &    0.637 &    0.411 &    0.237 &    0.270\\
                 &     1.3B &    0.400 &    0.670 &    0.460 &    0.247 &    0.298\\
                 &     2.7B &    0.493 &    0.704 &    0.495 &    0.287 &    0.332\\\thickhline
        \end{tabular}
    \end{minipage}
\end{figure}

\begin{figure}[p]
    \centering
    \begin{minipage}{0.9\linewidth}
        \centering
        \includegraphics[width=0.9\linewidth]{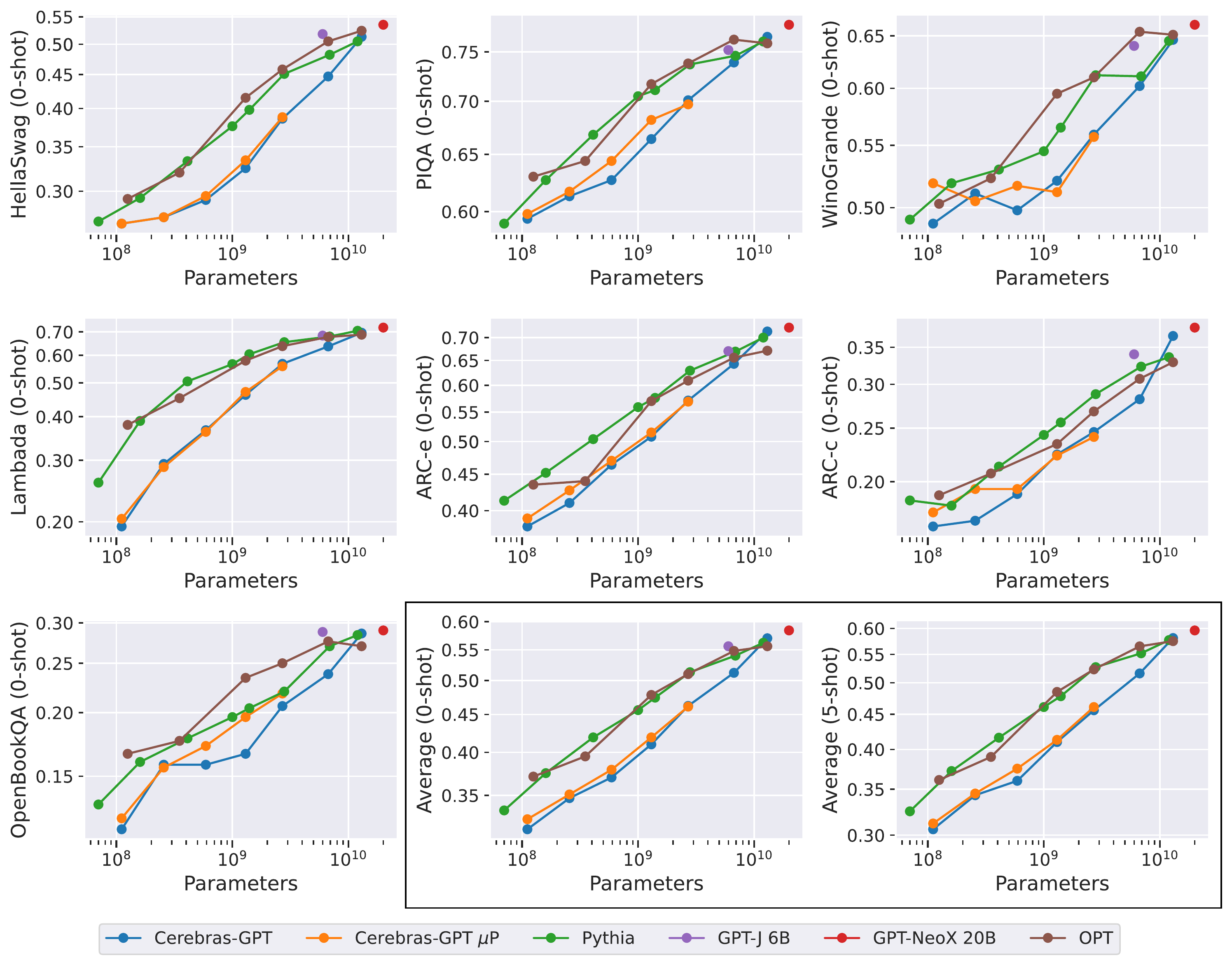}
        \vspace{-6pt}
        \caption{\centering Individual zero-shot downstream task accuracy, zero-shot average, and five-shot average, plotted against model parameters.}
        \label{figure:downstream_all_params}
    \end{minipage} \\
    \vspace{18pt}
    \begin{minipage}{0.9\linewidth}
        \centering
        \captionof{table}{Five-shot downstream task accuracy using length-normalized NLL selection criteria.}
        \vspace{-6pt}
        \label{table:5_shot_acc_norm}
        \begin{tabular}{lr|ccccc}\thickhline
        \multicolumn{2}{c|}{Model} & HellaSwag &     PIQA &    ARC-e &    ARC-c & OpenBookQA\\\hline
        \multirow{7}{*}{Cerebras-GPT} &     111M &    0.270 &    0.582 &    0.350 &    0.208 &    0.252\\
                 &     256M &    0.291 &    0.607 &    0.391 &    0.211 &    0.266\\
                 &     590M &    0.324 &    0.622 &    0.449 &    0.229 &    0.270\\
                 &     1.3B &    0.387 &    0.665 &    0.512 &    0.266 &    0.286\\
                 &     2.7B &    0.488 &    0.699 &    0.576 &    0.285 &    0.326\\
                 &     6.7B &    0.589 &    0.744 &    0.668 &    0.343 &    0.370\\
                 &      13B &    0.694 &    0.774 &    0.747 &    0.433 &    0.420\\\hline
        \multirow{5}{*}{Cerebras-GPT + \textmu P} &     111M &    0.275 &    0.583 &    0.343 &    0.202 &    0.268\\
                 &     256M &    0.288 &    0.614 &    0.391 &    0.225 &    0.262\\
                 &     590M &    0.337 &    0.646 &    0.449 &    0.227 &    0.270\\
                 &     1.3B &    0.401 &    0.670 &    0.516 &    0.263 &    0.294\\
                 &     2.7B &    0.492 &    0.701 &    0.591 &    0.294 &    0.322\\\thickhline
        \end{tabular}
    \end{minipage}
\end{figure}

For completeness, we include all downstream task results we collected for this study. Table \ref{table:eval_results_zero_shot} includes upstream Pile evaluations and all downstream zero-shot tasks for models GPT-J, GPT-NeoX, OPT, and Pythia, as well as Cerebras-GPT. Similarly, Table \ref{table:fewshot} shows the few-shot (five-shot) results for all models. Full downstream results are plotted in Figures \ref{figure:downstream_all_flops} and \ref{figure:downstream_all_params}.

Some prior works also use a different methods to select model predictions when evaluating the model's accuracy on some downstream tasks. Specifically, there are two commonly used techniques to select a model's prediction. First, the model can predict the probability of an output (continuation) sequence given a context sequence. Here, the selection criteria would be to select the continuation with maximum probability. We use this maximum probability approach in all prior results in the paper to be consistent with results in the GPT-NeoX paper. The second approach is to normalize the model's predicted probability in the log domain by the length of the continuation, and choose the continuation with the smallest length-normalized negative log-likelihood (NLL) ($\text{argmin}_i (-ln(p_i)/|c_i|)$, where $p_i$ is the model's predicted probability of continuation sequence $c_i$, and $|c_i|$ is the length of that sequence). This approach will tend to favor longer continuations with moderate probability, which might be preferred for some tasks. For comparison against prior works that report minimum length-normalized NLL, we report Cerebras-GPT results in Tables \ref{table:0_shot_acc_norm} and \ref{table:5_shot_acc_norm}.

\subsection{Differences Between Cerebras-GPT and Other Models}

The Cerebras-GPT 13B model improves over other publicly-available models of comparable size. This is a surprising result given that the creators of these other models modified the original GPT-2/3 architecture intending to improve convergence and training efficiency. There are many confounders that could contribute to Cerebras-GPT's advantages, but we briefly list known differences here to give an idea of the space of possible opportunities for future study.
\begin{itemize}
    \setlength\itemsep{-3px}
    \item GPT-J, GPT-NeoX, and Pythia models use rotary positional embeddings, which show modest loss/accuracy improvements and ability to extend to longer sequence lengths \citep{su2022roformer}. Cerebras-GPT uses standard trainable positional embeddings.
    \item Some GPT-J variants disable bias weights for fully-connected layers in transformer attention blocks. Other studies explain that disabling biases can increase accelerator utilization without loss degradation \citep{Chowdhery2022PaLMSL,vit22b}. We believe this approach might also improve training stability issues caused by key projection bias weights growth as we describe in Appendix \ref{appendix:stable_training}. We have not tested the effects on loss/accuracy from disabling these bias weights.
    \item GPT-J, GPT-NeoX, and Pythia use a parallel structure for attention and feed forward layers \citep{black2022gptneox}. This residual architecture has been reported to cause degradation in model performance at similar scales of models, so it is typically only adopted to increase accelerator utilization \citep{Chowdhery2022PaLMSL}. OPT and Cerebras-GPT models use the standard GPT-2 transformer block, which orders attention sequentially before the feed forward layers.
    \item GPT-NeoX and Pythia models use vocabulary and tokenization designed specifically for the Pile dataset \citep{black2022gptneox}. The resulting vocabulary is different in a few ways from the GPT-2/3 vocabulary. GPT-J, OPT, and Cerebras-GPT models use the GPT-2/3 vocabulary and tokenizer.
    \item Pythia models also include those that were trained on a deduplicated version of Pile \citep{eleuther2023pythia}. These models show an average of 1.2\% advantage on downstream tasks. This indicates further opportunity to improve models with further dataset curation.
    \item OPT are trained on a dataset combining the datasets used for RoBERTa, the PushShift.io Reddit dataset, and Pile, along with their own dataset pre-processing \citep{zhang2022opt}.
\end{itemize}

\begin{table}[p]
\caption{Pile pre-training test loss and zero-shot downstream task results for publicly available models.}
\vspace{-6pt}
\label{table:eval_results_zero_shot}
\resizebox{\textwidth}{!}
{
\begin{tabular}{cr|C{1.3cm}C{1.4cm}|C{1.3cm}cC{1.3cm}cccC{1.3cm}C{1.3cm}}
    \thickhline
        &    & \multicolumn{2}{c|}{Pre-training ($\downarrow$)} & \multicolumn{8}{c}{Downstream task accuracy ($\uparrow$)} \\
    \multicolumn{2}{c|}{Model} & Training FLOPs & Pile test xent & Hella- Swag &     PIQA & Wino- Grande &  Lambada &    ARC-e &    ARC-c & OpenBookQA & Downstream Average \\
   \hline
   GPT-J &     6.1B &  1.7e22 &    1.613 &    0.518 &    0.752 &    0.640 &    0.683 &    0.670 &    0.340 &    0.288 &    0.556\\
   \hline
   GPT-NeoX &      20B &  6.4e22 &    1.519 &    0.535 &    0.779 &    0.661 &    0.720 &    0.723 &    0.380 &    0.290 &    0.584\\
   \hline
\multirow{6}{*}{OPT} &     125M &  4.1e20 &        - &    0.292 &    0.630 &    0.503 &    0.379 &    0.435 &    0.189 &    0.166 &    0.371\\
         &     350M &  1.1e21 &        - &    0.320 &    0.644 &    0.523 &    0.452 &    0.440 &    0.207 &    0.176 &    0.395\\
         &     1.3B &  3.2e21 &        - &    0.415 &    0.717 &    0.595 &    0.579 &    0.570 &    0.234 &    0.234 &    0.478\\
         &     2.7B &  6.1e21 &        - &    0.458 &    0.738 &    0.610 &    0.637 &    0.609 &    0.268 &    0.250 &    0.510\\
         &     6.7B &  1.4e22 &        - &    0.505 &    0.763 &    0.654 &    0.677 &    0.656 &    0.307 &    0.276 &    0.548\\
         &      13B &  2.7e22 &        - &    0.524 &    0.759 &    0.651 &    0.687 &    0.671 &    0.329 &    0.270 &    0.556\\
         \hline
\multirow{8}{*}{Pythia} &      70M &  1.6e20 &    2.504 &    0.270 &    0.590 &    0.491 &    0.259 &    0.413 &    0.185 &    0.132 &    0.334\\
         &     160M &  4.1e20 &    2.186 &    0.293 &    0.627 &    0.519 &    0.389 &    0.452 &    0.181 &    0.160 &    0.375\\
         &     410M &  1.1e21 &    1.971 &    0.333 &    0.668 &    0.530 &    0.505 &    0.504 &    0.213 &    0.178 &    0.419\\
         &       1B &  2.2e21 &    1.845 &    0.376 &    0.705 &    0.545 &    0.566 &    0.559 &    0.243 &    0.196 &    0.456\\
         &     1.4B &  3.2e21 &    1.793 &    0.398 &    0.711 &    0.565 &    0.604 &    0.576 &    0.256 &    0.204 &    0.474\\
         &     2.8B &  6.1e21 &    1.720 &    0.451 &    0.737 &    0.612 &    0.654 &    0.629 &    0.288 &    0.220 &    0.513\\
         &     6.9B &  1.4e22 &    1.626 &    0.482 &    0.746 &    0.611 &    0.679 &    0.669 &    0.323 &    0.270 &    0.540\\
         &      12B &  2.4e22 &    1.582 &    0.505 &    0.761 &    0.645 &    0.705 &    0.700 &    0.336 &    0.284 &    0.562\\
         \hline
\multirow{8}{*}{\begin{tabular}[x]{@{}c@{}}Pythia\\Pile-dedup\end{tabular}} &      70M &  1.6e20 &    2.549 &    0.273 &    0.607 &    0.526 &    0.257 &    0.404 &    0.175 &    0.136 &    0.340\\
         &     160M &  4.1e20 &    2.204 &    0.294 &    0.632 &    0.509 &    0.370 &    0.451 &    0.204 &    0.172 &    0.376\\
         &     410M &  1.1e21 &    1.989 &    0.341 &    0.668 &    0.534 &    0.514 &    0.519 &    0.206 &    0.180 &    0.423\\
         &       1B &  2.2e21 &    1.858 &    0.387 &    0.712 &    0.546 &    0.585 &    0.568 &    0.241 &    0.212 &    0.464\\
         &     1.4B &  3.2e21 &    1.889 &    0.403 &    0.729 &    0.561 &    0.610 &    0.582 &    0.265 &    0.198 &    0.478\\
         &     2.8B &  6.1e21 &    1.724 &    0.466 &    0.743 &    0.612 &    0.672 &    0.662 &    0.299 &    0.232 &    0.526\\
         &     6.9B &  1.4e22 &    1.644 &    0.488 &    0.756 &    0.636 &    0.695 &    0.667 &    0.320 &    0.252 &    0.545\\
         &      12B &  2.4e22 &    1.601 &    0.516 &    0.761 &    0.639 &    0.712 &    0.697 &    0.341 &    0.280 &    0.564\\
         \hline
\multirow{7}{*}{Cerebras-GPT} &     111M &  2.6e18 &    2.608 &    0.268 &    0.594 &    0.488 &    0.194 &    0.380 &    0.166 &    0.118 &    0.315\\
         &     256M &  1.3e19 &    2.349 &    0.274 &    0.613 &    0.511 &    0.293 &    0.410 &    0.170 &    0.158 &    0.347\\
         &     590M &  6.1e19 &    2.181 &    0.291 &    0.627 &    0.498 &    0.366 &    0.464 &    0.190 &    0.158 &    0.370\\
         &     1.3B &  2.8e20 &    1.997 &    0.325 &    0.664 &    0.521 &    0.462 &    0.508 &    0.224 &    0.166 &    0.410\\
         &     2.7B &  1.1e21 &    1.834 &    0.386 &    0.701 &    0.559 &    0.567 &    0.571 &    0.246 &    0.206 &    0.462\\
         &     6.7B &  6.3e21 &    1.704 &    0.447 &    0.739 &    0.602 &    0.636 &    0.643 &    0.282 &    0.238 &    0.512\\
         &      13B &  2.3e22 &    1.572 &    0.513 &    0.766 &    0.646 &    0.696 &    0.714 &    0.367 &    0.286 &    0.570\\
         \hline
\multirow{5}{*}{\begin{tabular}[x]{@{}c@{}}Cerebras-GPT\\+ \textmu P\end{tabular}} &     111M &  2.6e18 &    2.588 &    0.268 &    0.598 &    0.519 &    0.204 &    0.390 &    0.176 &    0.124 &    0.325\\
         &     256M &  1.3e19 &    2.359 &    0.274 &    0.617 &    0.505 &    0.287 &    0.427 &    0.194 &    0.156 &    0.351\\
         &     590M &  6.1e19 &    2.155 &    0.295 &    0.644 &    0.517 &    0.362 &    0.470 &    0.194 &    0.172 &    0.379\\
         &     1.3B &  2.8e20 &    1.984 &    0.334 &    0.682 &    0.512 &    0.471 &    0.515 &    0.223 &    0.196 &    0.419\\
         &     2.7B &  1.1e21 &    1.846 &    0.388 &    0.697 &    0.557 &    0.558 &    0.569 &    0.241 &    0.218 &    0.461\\
         \thickhline
\end{tabular}
}
\end{table}

\begin{table}[p]
\centering
\caption{Five-shot downstream task accuracy results. Higher accuracy is better.}
\vspace{-6pt}
\label{table:fewshot}
\resizebox{\textwidth}{!}
{
\begin{tabular}{cc|C{1.3cm}cC{1.3cm}cccC{1.3cm}C{1.3cm}}
    \thickhline
    \multicolumn{2}{c|}{Model} & Hella- Swag &     PIQA & Wino-Grande &  Lambada &    ARC-e &    ARC-c & OpenBookQA & Downstream Avg. \\
    \hline
    GPT-J &     6.1B &    0.494 &    0.756 &    0.660 &    0.662 &    0.705 &    0.360 &    0.310 & 0.564 \\
    \hline
    GPT-NeoX &      20B &    0.538 &    0.774 &    0.683 &    0.698 &    0.746 &    0.410 &    0.326 & 0.596 \\
    \hline
    \multirow{6}{*}{OPT} &     125M &    0.289 &    0.628 &    0.520 &    0.303 &    0.426 &    0.197 &    0.166 & 0.361 \\
         &     350M &    0.321 &    0.647 &    0.521 &    0.384 &    0.464 &    0.208 &    0.184 & 0.390 \\
         &     1.3B &    0.413 &    0.726 &    0.597 &    0.553 &    0.604 &    0.273 &    0.230 & 0.485 \\
         &     2.7B &    0.458 &    0.749 &    0.616 &    0.603 &    0.651 &    0.305 &    0.276 & 0.523 \\
         &     6.7B &    0.505 &    0.773 &    0.663 &    0.660 &    0.692 &    0.340 &    0.318 & 0.565 \\
         &      13B &    0.524 &    0.763 &    0.684 &    0.678 &    0.714 &    0.358 &    0.306 & 0.575 \\
         \hline
    \multirow{8}{*}{Pythia} &      70M &    0.269 &    0.589 &    0.491 &    0.192 &    0.399 &    0.184 &    0.148 & 0.325 \\
         &     160M &    0.292 &    0.631 &    0.515 &    0.329 &    0.469 &    0.205 &    0.164 & 0.372 \\
         &     410M &    0.333 &    0.669 &    0.522 &    0.448 &    0.526 &    0.229 &    0.188 & 0.416 \\
         &       1B &    0.374 &    0.709 &    0.562 &    0.514 &    0.596 &    0.265 &    0.206 & 0.461 \\
         &     1.4B &    0.398 &    0.712 &    0.573 &    0.553 &    0.622 &    0.274 &    0.214 & 0.478 \\
         &     2.8B &    0.448 &    0.738 &    0.621 &    0.629 &    0.673 &    0.328 &    0.254 & 0.527 \\
         &     6.9B &    0.478 &    0.750 &    0.646 &    0.641 &    0.699 &    0.355 &    0.296 & 0.552 \\
         &      12B &    0.506 &    0.759 &    0.662 &    0.673 &    0.731 &    0.383 &    0.322 & 0.577 \\
    \hline
    \multirow{8}{*}{\begin{tabular}[x]{@{}c@{}}Pythia\\Pile-dedup\end{tabular}} &      70M &    0.272 &    0.604 &    0.519 &    0.192 &    0.403 &    0.177 &    0.152 & 0.331 \\
         &     160M &    0.294 &    0.639 &    0.507 &    0.309 &    0.472 &    0.215 &    0.178 & 0.373 \\
         &     410M &    0.339 &    0.673 &    0.513 &    0.456 &    0.537 &    0.232 &    0.190 & 0.420 \\
         &       1B &    0.384 &    0.710 &    0.552 &    0.529 &    0.588 &    0.259 &    0.226 & 0.464 \\
         &     1.4B &    0.400 &    0.730 &    0.566 &    0.565 &    0.617 &    0.283 &    0.232 & 0.485 \\
         &     2.8B &    0.463 &    0.758 &    0.609 &    0.637 &    0.681 &    0.327 &    0.282 & 0.537 \\
         &     6.9B &    0.492 &    0.762 &    0.637 &    0.671 &    0.705 &    0.344 &    0.308 & 0.560 \\
         &      12B &    0.516 &    0.765 &    0.678 &    0.696 &    0.728 &    0.386 &    0.326 & 0.585 \\
         \hline
    \multirow{7}{*}{Cerebras-GPT} &     111M &    0.267 &    0.588 &    0.475 &    0.158 &    0.356 &    0.166 &    0.136 & 0.306 \\
         &     256M &    0.278 &    0.606 &    0.522 &    0.225 &    0.422 &    0.183 &    0.164 & 0.343 \\
         &     590M &    0.291 &    0.634 &    0.479 &    0.281 &    0.475 &    0.206 &    0.152 & 0.360 \\
         &     1.3B &    0.326 &    0.668 &    0.536 &    0.395 &    0.529 &    0.241 &    0.174 & 0.410 \\
         &     2.7B &    0.382 &    0.697 &    0.543 &    0.487 &    0.590 &    0.267 &    0.224 & 0.456 \\
         &     6.7B &    0.444 &    0.736 &    0.590 &    0.591 &    0.667 &    0.314 &    0.270 & 0.516 \\
         &      13B &    0.514 &    0.768 &    0.674 &    0.655 &    0.743 &    0.398 &    0.318 & 0.581 \\
    \hline
    \multirow{5}{*}{\begin{tabular}[x]{@{}c@{}}Cerebras-GPT\\+ \textmu P\end{tabular}} &     111M &    0.268 &    0.581 &    0.520 &    0.146 &    0.368 &    0.175 &    0.124 & 0.312 \\
         &     256M &    0.278 &    0.619 &    0.534 &    0.220 &    0.415 &    0.193 &    0.154 & 0.345 \\
         &     590M &    0.298 &    0.652 &    0.515 &    0.301 &    0.479 &    0.206 &    0.174 & 0.375 \\
         &     1.3B &    0.329 &    0.672 &    0.513 &    0.396 &    0.531 &    0.235 &    0.212 & 0.413 \\
         &     2.7B &    0.382 &    0.704 &    0.560 &    0.510 &    0.595 &    0.267 &    0.210 & 0.461 \\
    \thickhline
\end{tabular}
}
\end{table}

\newpage

\subsection{Bias}
\label{section:bias}

Language models carry with them the risk of causing harm through the propagation of bias, toxicity, and other negative traits found in their training data. Accordingly, it is important to test models for such biases. We evaluate our models on the CrowS-Pairs dataset \citep{crowspairs}, which measures bias across nine different categories. In Table \ref{table:bias}, we compare bias measurements for our family of models to Pythia 70M--12B, as well as three well regarded baselines: GPT-3 175B \citep{gpt3}, OPT 175B \citep{zhang2022opt}, and LLaMA 65B \citep{llama}.

\begin{table}[h]
\centering
\caption{\centering Analyzing levels of bias using the CrowS-Pairs dataset, comparing GPT-3 175B, OPT 175B, Pythia, and LLaMA 65B models to Cerebras-GPT. Higher values correspond to higher bias.}
\vspace{-6pt}
\resizebox{\textwidth}{!}{
\begin{tabular}{lr|C{1.1cm}C{1.3cm}C{1.2cm}C{1.1cm}C{1.3cm}C{1.3cm}C{1.3cm}C{1.3cm}C{1.3cm}|C{1.3cm}}\thickhline
    \multicolumn{2}{c|}{Model} & Race/ Color & Socioeconomic status &   Gender &      Age & Religion & Disability & Sexual orientation & Nationality & Physical appearance &  Average \\
    \hline
    GPT-3 &     175B &     64.7 &     73.8 &     62.6 &     64.4 &     73.3 &     76.7 &     76.2 &     61.6 &     74.6 &     67.2 \\
    \hline
     OPT &     175B &     68.6 &     76.2 &     65.7 &     67.8 &     68.6 &     76.7 &     78.6 &     62.9 &     76.2 &     69.5 \\
    \hline
    \multirow{8}{*}{Pythia} &      70M &     51.2 &     63.7 &     58.1 &     54.9 &     64.0 &     66.2 &     79.6 &     41.2 &     66.7 &     56.6 \\
         &     160M &     49.8 &     56.8 &     55.9 &     56.0 &     72.1 &     66.2 &     73.1 &     45.4 &     63.9 &     55.8 \\
         &     410M &     53.7 &     62.6 &     61.9 &     63.7 &     65.8 &     72.3 &     81.7 &     52.3 &     62.5 &     60.1 \\
         &       1B &     54.1 &     65.8 &     62.2 &     63.7 &     72.1 &     73.8 &     78.5 &     52.3 &     63.9 &     61.1 \\
         &     1.4B &     51.8 &     65.8 &     63.4 &     62.6 &     76.6 &     72.3 &     82.8 &     57.4 &     68.1 &     61.8 \\
         &     2.8B &     53.7 &     66.3 &     63.1 &     63.7 &     78.4 &     78.5 &     83.9 &     55.1 &     73.6 &     62.9 \\
         &     6.9B &     55.5 &     72.6 &     66.6 &     72.5 &     80.2 &     72.3 &     84.9 &     56.5 &     76.4 &     65.6 \\
         &      12B &     55.9 &     68.4 &     63.4 &     68.1 &     75.7 &     72.3 &     83.9 &     57.9 &     73.6 &     64.0 \\
    \hline
    LLaMA &      65B &     57.0 &     71.5 &     70.6 &     70.1 &     79.0 &     66.7 &     81.0 &     64.2 &     77.8 &     66.6 \\
    \hline
    \multirow{7}{*}{Cerebras-GPT} &     111M &     41.3 &     69.5 &     55.6 &     42.9 &     64.9 &     60.0 &     78.5 &     43.5 &     61.1 &     52.9 \\
         &     256M &     52.8 &     63.2 &     57.8 &     53.8 &     60.4 &     67.7 &     80.6 &     44.4 &     61.1 &     56.9 \\
         &     590M &     46.9 &     62.6 &     58.1 &     59.3 &     79.3 &     64.6 &     79.6 &     47.7 &     66.7 &     57.2 \\
         &     1.3B &     50.6 &     60.5 &     58.1 &     61.5 &     73.0 &     69.2 &     73.1 &     45.4 &     72.2 &     57.6 \\
         &     2.7B &     53.7 &     65.8 &     60.3 &     64.8 &     76.6 &     67.7 &     78.5 &     52.8 &     69.4 &     61.1 \\
         &     6.7B &     54.1 &     65.3 &     64.4 &     65.9 &     80.2 &     72.3 &     86.0 &     59.7 &     73.6 &     63.9 \\
         &      13B &     55.1 &     72.1 &     67.5 &     73.6 &     81.1 &     73.8 &     78.5 &     59.7 &     75.0 &     65.7 \\
    \thickhline
\end{tabular}}
\label{table:bias}
\end{table}

The Cerebras-GPT models exhibit less bias on average than any of the larger model baselines. However, Cerebras-GPT 13B does show bias greater than GPT-3, OPT, or LLaMa on six of the nine bias categories, indicating that compute-efficient pre-training is not immune to large bias.

When observing bias levels across Cerebras-GPT or Pythia models, we see that biases tend to grow with model size. The Cerebras-GPT models tend to show a larger range of bias values over the growing model sizes (e.g., gender), while Pythia models sometimes show similar bias across model sizes (e.g., disability). This suggests that models trained on a fixed dataset size may be likely to extract similar levels of bias regardless of model size. On the other hand, more compute-efficient training (smaller datasets for smaller models) might mitigate some bias issues compared to models trained on more data.

Finally, we note that when comparing Cerebras-GPT and Pythia models trained with similar compute budgets, Pythia models tend to have slightly lower bias. In particular, the Cerebras-GPT models 1.3B, 2.7B, 6.7B, and 13B use similar compute to Pythia models 160M, 410M, 2.8B, and 12B, respectively. These Cerebras-GPT models show roughly 1-2\% higher bias on average. This suggests that bias is more efficiently extracted from the training data when using a compute-optimal pre-training setup, possibly due to the larger model sizes.

Overall, we recommend further bias evaluation and mitigations for Cerebras-GPT and larger models if deploying them in production settings.

\section{Additional Tokens-per-parameter Experiments}
\label{appendix:tokens_per_parameter}

Here, we give more evidence that 20 tokens-per-parameter is nearly compute-optimal when pre-training GPT-like models on the Pile dataset.

\subsection*{Estimated Chinchilla Losses}
First, in Figure \ref{figure:tpp} in Section \ref{section-results}, we include a curve that estimates the Chinchilla loss degradation when changing the number of tokens per parameter for which a model is pre-trained. To get that estimate, we start by fitting a curve to points we estimate in the Chinchilla paper plot (Figure 3), which shows loss for different model sizes and tokens trained with fixed compute budgets. Given our approach, our estimates are likely to introduce error, and we do not know the true functional form of their fits. However, we pull points from three different FLOP levels, and we validate our curve fit has low error for a fourth held-out FLOP level. This result seems surprising that large changes in model size and training tokens do not appear to have a large effect on the expected degradation from computationally-inefficient training, but it might indicate another invariant to training scale.

We use the Chinchilla curve fit to estimate the proportional loss degradation, $\Delta\mathcal{L}$, when changing tokens-per-parameter, $\tau$ (this is the Chinchilla trend plotted in Figure \ref{figure:tpp}):

\vspace{-6pt}
\begin{equation}
    \Delta\mathcal{L}(\tau) = 0.023 \cdot ln(\sqrt{20/\tau})^2
\end{equation}

This degradation formulation shows good agreement with our tests. Further, Chinchilla models were trained on MassiveText, while our models and Pythia models were trained on the Pile, suggesting the two datasets have significant commonality in their scaling characteristics when training on more than 20 tokens per parameter.

\subsection*{Our Tokens-per-parameter Experiments}

\begin{figure}[h]
    \centering
    \begin{subfigure}[b]{0.49\textwidth}
        \centering
        \includegraphics[width=\linewidth]{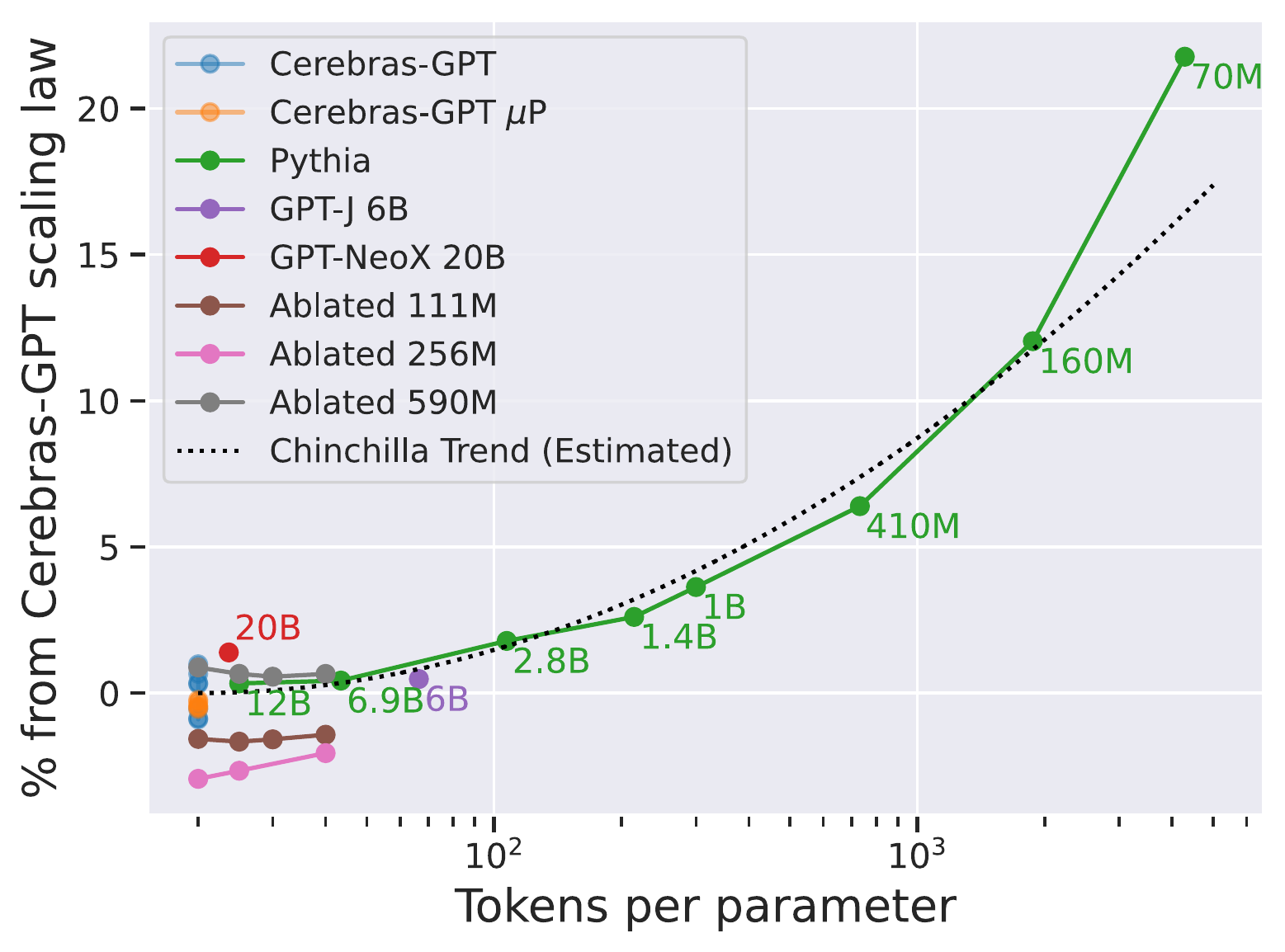}
    \end{subfigure}
    \hfill
    \centering
    \begin{subfigure}[b]{0.49\textwidth}
        \centering
        \includegraphics[width=\linewidth]{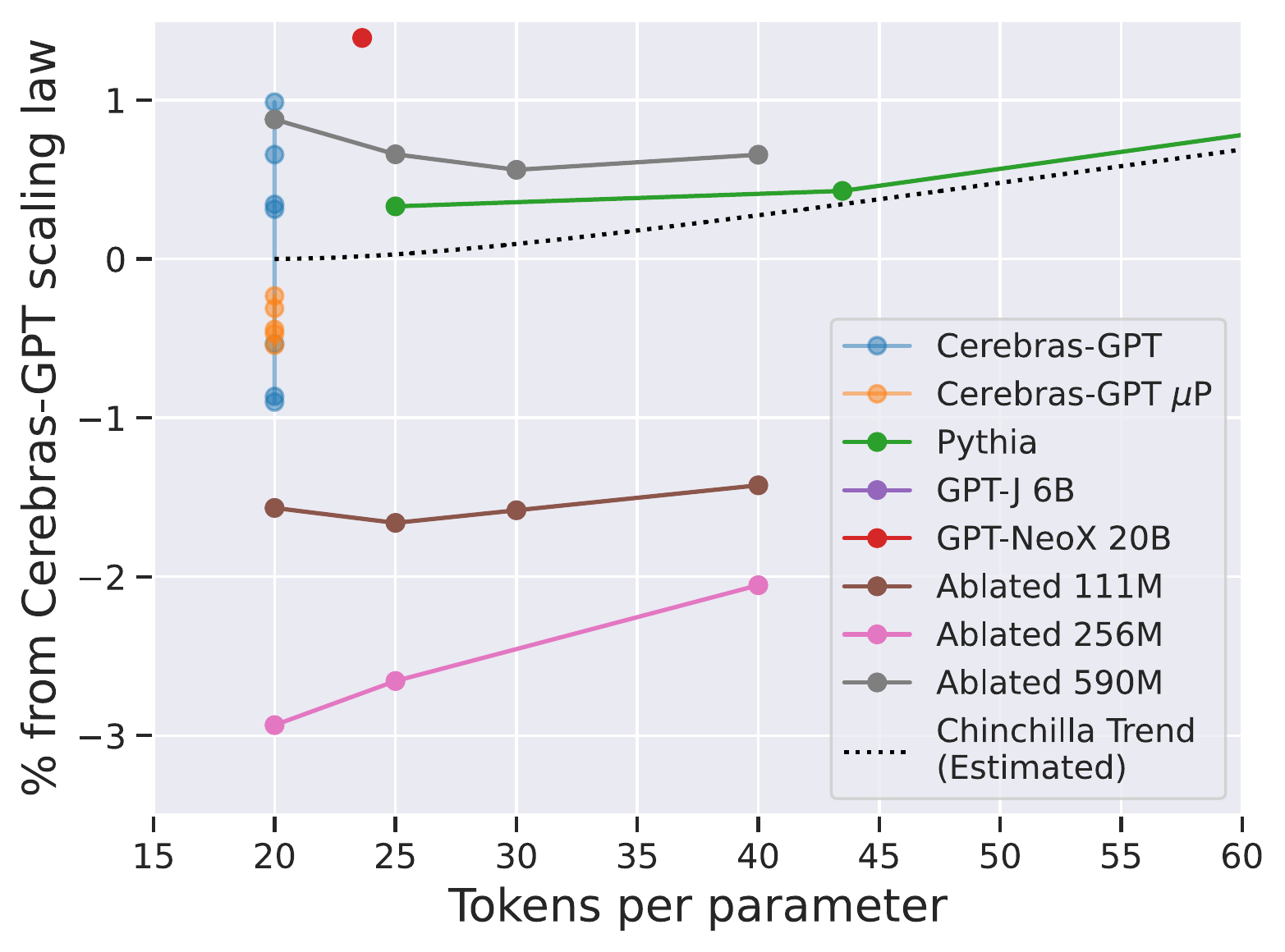}
    \end{subfigure}
    \vspace{-6pt}
    \caption{\centering Percent loss degradation compared to the Cerebras-GPT compute-efficient scaling law for varying tokens per parameter.}
    \label{figure:tpp_detail_tests}
\end{figure}

Figure \ref{figure:tpp_detail_tests} (left) plots the loss degradation (\%) from our Cerebras-GPT compute-efficient scaling law for different tokens per parameter (similar to Figure \ref{figure:tpp}), and we add our small scale experiments around 20 tokens per parameter for model sizes 111M, 256M, and 590M parameters. The right plot in Figure \ref{figure:tpp_detail_tests} zooms in on the region from 15 to 50 tokens per parameter. The right plot shows that losses are quite stable between 20 and 40 tokens per parameter and our empirical best results are between 20 and 30 for each of the three models. We note that the variance in loss for runs at this scale is roughly 0.35\%, so most loss values here are also within expected run-to-run variance. Based on these results and the strong agreement between Chinchilla and Pythia results, we were comfortable to conclude that 20 tokens-per-parameter is nearly compute-optimal for these models trained the Pile dataset.

\section{Number of Model Parameters}
\label{appendix:n_params}

Table \ref{code:params_count} shows the formula we use to calculate parameter counts for Cerebras-GPT models.

\begin{table}[h]
\caption{Python code to calculate parameter count in Cerebras-GPT models}
\vspace{-6pt}
\label{code:params_count}
{\footnotesize
\begin{python}
def get_n_params(vocab_size, d_model, num_layers, seq_length=2048):
    embedding = vocab_size * d_model + d_model * seq_length
    ln1 = 2 * d_model
    attn = 4 * (d_model**2 + d_model)
    ln2 = 2 * d_model
    ffn = 8 * d_model**2 + 5 * d_model
    encoder = num_layers * (ln1 + attn + ln2 + ffn)
    final_ln = 2 * d_model
    n_params = embedding + encoder + final_ln
    return n_params
\end{python}
}
\end{table}

\section{Number of Training FLOPs}
We calculate the number of training FLOPs with a formula similar to Chinchilla, but with two modifications. First, we account for the dot product between $softmax(QK^T)$ and $V$. Second, we account for the fact that embedding layers do not need to calculate a delta gradient for earlier layers. Code for FLOPs count is in Table \ref{code:flops_count}. We consider this FLOP calculation to be a measure of the algorithmic calculations required for forward and backward gradient steps of training, or ``Algorithmic FLOPs''. This formula does not include any additional FLOPs for things like activation checkpointing and recomputation or specifics related to software or hardware implementation.

\begin{table}[h]
\caption{Python code to calculate FLOPs per sequence in Cerebras-GPT models}
\vspace{-6pt}
\label{code:flops_count}
{\footnotesize
\begin{python}
def get_flops_per_seq(vocab_size, d_model, num_layers, key_size, seq_len, inference=False):
    num_heads = d_model // key_size
    embeddings = 2 * seq_len * vocab_size * d_model
    position_embeddings = 2 * d_model * seq_len

    kqv_proj = num_layers * 2 * 3 * seq_len * d_model * (key_size * num_heads)
    kq_logits = num_layers * 2 * (seq_len**2) * (key_size * num_heads)
    softmax = num_layers * 3 * (key_size * num_heads) * (seq_len**2)
    softmax_q_red = num_layers * (seq_len**2) * (key_size * num_heads)
    final_linear = num_layers * 2 * seq_len * (key_size * num_heads) * d_model
    sm_v_dot = num_layers * 2 * (seq_len**2) * (key_size * num_heads)
    attention_blocks = kqv_proj + kq_logits + softmax + softmax_q_red + sm_v_dot + final_linear

    dense_blocks = num_layers * 16 * seq_len * (d_model**2)
    final_logits = 2 * seq_len * d_model * vocab_size

    # Layer norms: 7 FLOPs/activation, 2 LNs per decoder block
    layer_norm_flops = num_layers * 2 * 7 * (seq_len * d_model)

    # GeLU: estimate 20 FLOPs/activation, applied in FFN with 4x hidden dim
    gelu_flops = num_layers * 20 * 4 * (seq_len * d_model)

    total_flops_per_step = embeddings + position_embeddings + layer_norm_flops + attention_blocks + dense_blocks + final_logits + gelu_flops
    inference_flops_per_step = total_flops_per_step

    # Account for backward pass too
    total_flops_per_step *= 3

    # Embeddings don't need to pass a delta back
    total_flops_per_step -= embeddings
    total_flops_per_step -= position_embeddings

    if inference:
        return inference_flops_per_step
    else:
        return total_flops_per_step
\end{python}
}
\end{table}

\newpage
\section{Additional \textmu P Details}
\label{appendix:mup_implementation}
We spent time orienting ourselves to use \textmu P, so this section describes that process for other practitioners.

\subsection{\textmu P Implementation}
In Table \ref{table:mup-implementation} we detail all the changes required to implement \textmu P in GPT-like models, and we describe these changes in more detail here.
\vspace{-6pt}
\begin{itemize}
    \setlength\itemsep{-2px}
    \item \textmu P adds a tunable embedding output activation multiplier, $m_{\text{emb}}$, which is multiplied by the sum of token and position embeddings. This multiplier controls relative activation and gradient magnitudes between the embeddings layers and the transformer backbone.
    \item Similarly, to control the the relative gradient magnitude between embeddings and the transformer backbone, the model's output logits activation tensor (pre-softmax) is multiplied by $1/m_{\text{width}}$ (when using shared embedding and output weights).
    \item To control the expected magnitude of activations in the transformer blocks, \textmu P scales the initialization variance for each fully-connected layer's weights by $1/m_{\text{width}}$.
    \item To control the relative weight magnitudes throughout training, \textmu P scales the learning rate of each fully-connected layer's weights by $1/m_{\text{width}}$.
    \item Under the assumption that query and key projections have significant alignment later in training, \textmu P scales the key-query dot-product activations by $1/d_{\text{head}}$, rather than using the $1/\sqrt{d_{\text{head}}}$ scaling originally proposed in \cite{transformer2017}.
    \item The next subsection describes how we tune the three transferrable \textmu P hyperparameters: the base learning rate ($\eta_{\text{base}}$), the base initialization standard deviation ($\sigma_{\text{base}}$), and the embedding output multiplier ($m_{\text{emb}}$).
\end{itemize}

{
\renewcommand{\arraystretch}{1.05}
\begin{table}[p]
    \centering
    \caption{Cheat Sheet: All implementation details required to compare SP and \textmu P.}
    \vspace{-6pt}
    \label{table:mup-implementation}
    \begin{tabular}{|l|c|c|}
        \multicolumn{1}{l}{}        & ~~Standard Parameterization (SP)~~ & \multicolumn{1}{c}{~~~~~~~~~Maximal Update (\textmu P)~~~~~~~~~} \\
        \hline
        \multicolumn{3}{l}{\bf Variables} \\        
        \hline \hline
        $W$                         & \multicolumn{2}{c|}{A multiplicative or fully-connected weights tensor} \\
        $b$                         & \multicolumn{2}{c|}{A bias weights tensor} \\
        $X$, $Y$                    & \multicolumn{2}{c|}{Activation tensors: layer input, output, respectively} \\
        $d_{\text{model,base}}$     & \multicolumn{2}{c|}{Proxy (base) model's layer width} \\
        $d_{\text{model}}$          & \multicolumn{2}{c|}{Width of each layer} \\
        $d_{\text{head}}$           & \multicolumn{2}{c|}{Size of each attention head} \\
        $\text{embed}$              & \multicolumn{2}{c|}{Combined token + position embedding function} \\
        $\eta_{\text{base}}$        & \multicolumn{2}{c|}{The base learning rate (LR): Maximum in training schedule} \\
        $\sigma_{\text{base}}$      & \multicolumn{2}{c|}{The base initialization standard deviation} \\
        $m_{\text{width}}$          & --- & Layer width multiplier: \\
                                    &     & $d_{\text{model}} / d_{\text{model,base}}$ \\
        $m_{\text{emb}}$            & --- & Emdedding output multiplier \\

        \hline
        \multicolumn{3}{l}{\bf Empirically Tuned Values} \\
        \hline \hline
        $d_{\text{model,base}}$ & --- & 256 \\
        $\eta_{\text{base}}$                  & Must tune for each model size  & $6e$-3 \\
        $\sigma_{\text{base}}$  & 0.02                           & $0.08$ \\
        $m_{\text{emb}}$        & ---                            & $10.0$ \\

        \hline
        \multicolumn{3}{l}{\bf Formulas} \\
        \hline \hline
        Embedding initializer       & \multicolumn{2}{c|}{$W_{\text{emb}} \sim N_{\text{trunc}}(0, \sigma_{\text{base}}^2)$} \\
        Embedding LR                & \multicolumn{2}{c|}{$\eta_{\text{emb}} = \eta_{\text{base}}$} \\
        Embedding output & $Y_\text{emb} = \text{embed}(X)$
                                & $Y_\text{emb} = m_{\text{emb}} \cdot \text{embed}(X)$ \\
        LN initializer          & \multicolumn{2}{c|}{$W_\gamma \sim 1, b_\beta \sim 0$} \\
        LN LR                   & \multicolumn{2}{c|}{$\eta_{\text{LN}} = \eta_{\text{base}}$} \\
        Bias initializer        & \multicolumn{2}{c|}{$b \sim 0$} \\
        Bias LR                 & \multicolumn{2}{c|}{$\eta_b = \eta_{\text{base}}$} \\
        MHA equation            & $\text{softmax}\left(\frac{Q^TK}{\sqrt{d_{\text{head}}}}\right)V$
                                & $\text{softmax}\left(\frac{Q^TK}{d_{\text{head}}}\right)V$ \\
        QKV weights initializer & $W_{\text{qkv}} \sim N_{\text{trunc}}(0, \sigma_{\text{base}}^2)$
                                & $W_{\text{qkv}} \sim N_{\text{trunc}}(0, \sigma_{\text{base}}^2 / m_{\text{width}})$ \\
        QKV weights LR          & $\eta_{\text{qkv}} = \eta_{\text{base}}$
                                & $\eta_{\text{qkv}} = \eta_{\text{base}} / m_{\text{width}}$ \\
        O weights initializer   & $W_{\text{o}} \sim N_{\text{trunc}}(0, \frac{ \sigma_{\text{base}}^2}{2 \cdot n_{\text{layers}}})$
                                & $W_{\text{o}} \sim N_{\text{trunc}}(0, \frac{ \sigma_{\text{base}}^2}{2m_{\text{width}} \cdot n_{\text{layers}}})$ \\
        O weights LR            & $\eta_{\text{o}} = \eta_{\text{base}}$
                                & $\eta_{\text{o}} = \eta_{\text{base}} / m_{\text{width}}$ \\
        ffn1 weights initializer & $W_{\text{ffn1}} \sim N_{\text{trunc}}(0, \sigma_{\text{base}}^2)$
                                 & $W_{\text{ffn1}} \sim N_{\text{trunc}}(0, \sigma_{\text{base}}^2/m_{\text{width}})$ \\
        ffn1 weights LR         & $\eta_{\text{ffn1}} = \eta_{\text{base}}$
                                & $\eta_{\text{ffn1}} = \eta_{\text{base}} / m_{\text{width}}$ \\
        ffn2 weights initializer & $W_{\text{ffn2}} \sim N_{\text{trunc}}(0, \frac{\sigma_{\text{base}}^2}{2 \cdot n_{\text{layers}} })$
                                 & $W_{\text{ffn2}} \sim N_{\text{trunc}}(0, \frac{\sigma_{\text{base}}^2}{2m_{\text{width}} \cdot n_{\text{layers}}})$ \\
        ffn2 weights LR         & $\eta_{\text{ffn2}} = \eta_{\text{base}}$
                                & $\eta_{\text{ffn2}} = \eta_{\text{base}} / m_{\text{width}}$ \\
        Output logits multiplier & $Y_\text{logits} = W_{unemb}X$
                                 & $Y_\text{logits} = W_{unemb}X / m_{\text{width}}$ \\
        \hline
    \end{tabular}
\end{table}
}

\subsection{\textmu P Hyperparameter Search Details}
\label{section:mup_hp}
\begin{figure}[h]
    \centering
    \includegraphics[width=1\linewidth]{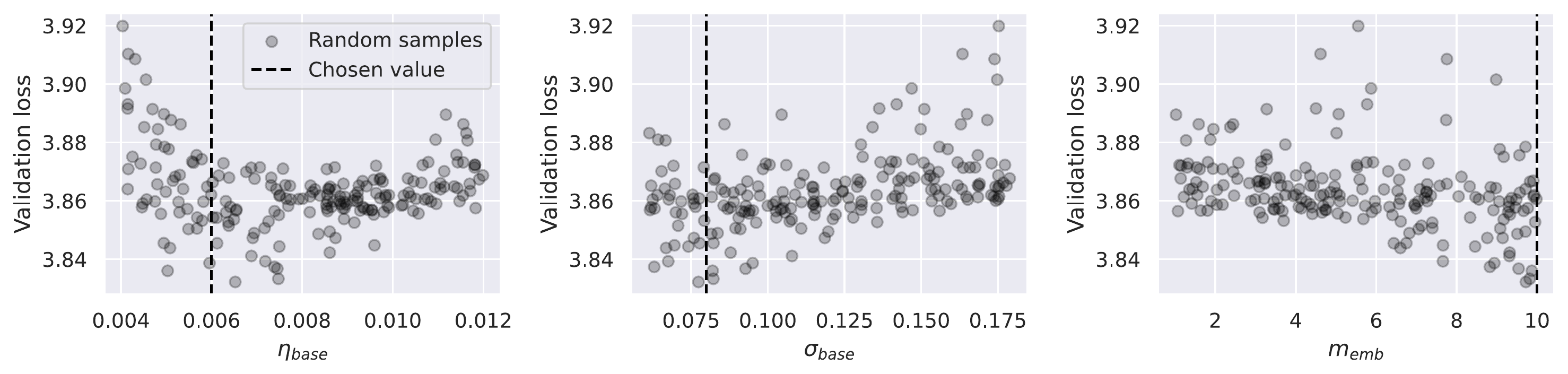}
    \vspace{-6pt}
    \caption{\centering Results from random hyperparameter search on 40M parameter \textmu P proxy model.}
    \label{figure:mup_hp_search}
\end{figure}

We tune \textmu P hyperparameters on a 40M parameter proxy model, and \textmu Transfer those hyperparameters to our models with 111M--2.7B parameters. Figure \ref{figure:mup_hp_search} shows the results of a 200 sample random hyperparameter search on a 40M parameter proxy model ($d_{\text{model}}=d_{\text{model,base}}=256$, $n_{\text{layers}}=32$, $d_{\text{head}}=128$) trained on 600M tokens with a batch size of 131k tokens. From this sweep we obtained the following tuned hyperparameters for our \textmu P models: $\eta_{\text{base}} = 6e$-$3$, $\sigma_{\text{base}} = 0.08$, $m_{\text{emb}} = 10$. These hyperparameters also closely match those used by \cite{yang2022mup}.

\subsection{Advice for Practitioners}

\subsubsection*{Use Large Enough Batch Sizes For Each Model Size}
The \textmu P paper suggests small proxy models should be trained with batch sizes similar to the larger target model to which you would like to \textmu Transfer the hyperparameters. We find, however, that choice of proxy model batch size can be more flexible as long as it is large enough. When tuning on a small model, Yang and Hu choose batch size to be quite large such that it is an appropriate batch size for the largest models to which they \textmu Transfer those hyperparameters. We find that proxy model tuning does {\it not} need to be performed on a batch size appropriate for the largest model. Rather, batch size must be sufficiently large such that the proxy model's gradient dynamics are likely to be consistent with the larger models. More specifically, setting the proxy model's batch size larger than its critical batch size (dictated by gradient noise \citep{mccandlish2018empirical}) is sufficient to get good hyperparameter transferrability. Then, batch size should be scaled appropriately as model size scales.

\subsubsection*{Simple Batch Size + Learning Rate Scaling with \textmu P}
More precisely, we find that \textmu P learning rate transfers as long as each model size is trained with a batch size roughly consistent with or larger than the critical batch size. The closer the batch size is to the critical batch size for a given model, the better the loss will be when using the \textmu Transferred learning rate. Further, when training models with a batch size smaller than the critical batch size, learning rate should be reduced linearly proportional to the reduction in batch size--consistent with the findings in \citep{google-batch-sizes, yang2022mup}.

\end{document}